\documentclass[10pt]{article}
\usepackage[english]{babel}
\usepackage{amssymb}
\usepackage{amsmath}
\usepackage{pdfpages}
\usepackage{caption}
\usepackage[letterpaper,top=2cm,bottom=2cm,left=3cm,right=3cm,marginparwidth=1.75cm]{geometry}

\usepackage{amsmath}
\usepackage{graphicx}
\usepackage[colorlinks=true, linkcolor=black, citecolor=blue]{hyperref}
\usepackage{lineno}
\usepackage{color}
\usepackage[superscript]{cite}
\usepackage{float}
\usepackage{graphicx} 
\usepackage{caption}  
\usepackage[normalem]{ulem}
\usepackage{xcolor}   
\usepackage{xspace}
\usepackage{caption}
\usepackage{listings}
\usepackage{tcolorbox}
\tcbuselibrary{listings, skins, breakable} 

\def\ourAG{\textit{SpineAgent}\xspace}
\def\ourFM{\textit{SpineFM}\xspace}

\newcounter{suppfigure}

\renewcommand{\thesuppfigure}{\arabic{suppfigure}} 

\newcommand{\suppfigurecaption}[1]{
  \refstepcounter{suppfigure}
  \caption*{\textbf{Supplementary Figure \thesuppfigure:} #1} 
}
\makeatletter
\renewcommand{\maketitle}{\bgroup\setlength{\parindent}{0pt}
\begin{flushleft}
  \textbf{\@title}

  \@author
\end{flushleft}\egroup
}
\makeatother

\newcommand{\beginToken}{<|begin\_of\_text|>}
\newcommand{\headerToken}[1]{<|start\_header\_id|#1|end\_header\_id|>}

\newcommand{\imageToken}{%
  \texttt{<image>}%
  \textsuperscript{\textnormal{%
    \hyperref[note:visual_input]{\textcolor{blue!70}{(\ref*{note:visual_input})}}%
  }}%
}

\newtcblisting{promptbox}[2][]{%
  colback=gray!2,
  colframe=gray!30!black,
  fonttitle=\bfseries, 
  title={#2},
  breakable,
  enhanced,
  listing only,
  listing options={
    basicstyle=\ttfamily\normalsize, 
    breaklines=true,
    columns=fullflexible,
    showstringspaces=false,
    escapeinside={(*}{*)},
  },
  #1
}

\newtcolorbox[auto counter]{inputnote}{
    colback=blue!5!white,
    colframe=blue!75!black,
    fonttitle=\bfseries,
    title={Visual Input Specification \hfill Note \thetcbcounter}, 
    fontupper=\ttfamily\normalsize, 
    arc=1mm,
    boxrule=0.5pt,
    label={note:visual_input},
}

\begin{document}
\title{A multi-agent system for spine MRI report generation from multi-sequence imaging}

\author{
Zhiping Xiao$^{1,*}$, Junwei Yang$^{2,*}$, Gongbo Sun$^{3,*}$, 
Han Zhang$^{1}$, Hanwen Xu$^{1}$, Yi Yao$^{4}$, 
Zachary D. Miller$^{5}$, William E. King III$^{5}$, Mohammed M. Kanani$^{6}$, Jalal B. Andre$^{5}$, Sammy Chu$^{5}$, 
Ming Zhang$^{2}$,
Paul E. Kinahan$^{5}$, Nathan M. Cross$^{5,\#}$, Sheng Wang$^{1,\#}$ \\
$^{1}$Paul G. Allen School of Computer Science \& Engineering, University of Washington, Seattle, WA, USA \\
$^{2}$School of Computer Science, Peking University, Beijing, China\\
$^{3}$Department of Computer Sciences, University of Wisconsin--Madison, Madison, WI, USA \\
$^{4}$Department of Arts and Sciences, New York University, New York City, NY, USA \\
$^{5}$Department of Radiology, University of Washington Medical Center, Seattle, WA, USA \\
$^{6}$School of Medicine, University of Washington, Seattle, WA, USA \\
$^{\#}$Corresponding author. Email: nmcross@uw.edu,swang@cs.washington.edu
}

\maketitle

\begin{abstract}
Spinal pathology is a leading cause of pain and disability worldwide. Spine magnetic resonance imaging (MRI) is central to clinical evaluation, yet its interpretation remains complex and time-consuming, requiring integration of information across multiple imaging sequences and anatomical regions. Despite recent advances in automated MRI analysis, effectively combining multi-sequence data while preserving sequence-specific diagnostic information remains an open challenge.

Here we present \ourAG, a multi-agent framework for spine MRI report generation built upon a multi-sequence foundation model trained on routine clinical data from $32{,}047$ patients and $453{,}683$ MRI series, comprising a total of $13{,}441{,}191$ MRI slices. To accommodate diverse modalities of sequences, we first pre-train two DINOv3-based encoders separately on T1- and T2-weighted sequences. We then introduce a continual training strategy that learns a synthesizer to embed images of other sequences using the T1 and T2 encoders, producing patient-level embedding that integrates various signals across MRI sequences. Using these embeddings, \ourAG achieves state-of-the-art performance, with mean $10.8\%$ AUROC improvement across $17$ spinal condition-prediction tasks compared to the best competing method, and demonstrates strong generalizability under cross-manufacturer and cross-cohort evaluation. Beyond classification, \ourAG enables pathology localization by identifying findings-relevant slices and segmenting pathological regions. It also supports multimodal image–report retrieval, providing a solid foundation for scalable and explainable MRI report generation.

We further integrate these validated capabilities of \ourAG into $37$ specialized agents for condition diagnosis, pathological-region localization, and clinically-similar-cases retrieval. Finally, we incorporate their outputs as structured tokens within a Medical Report Agent trained end-to-end for report generation. Through both automated metrics and expert evaluation by five radiologists, \ourAG achieves leading performance in spine MRI report generation.

Together, \ourAG introduces a continual training approach for multi-sequence spine MRI understanding. By decomposing report generation into clinically grounded subtasks addressed by specialized agents, the \ourAG framework enables accurate, interpretable and generalizable spine MRI reporting across diverse imaging sequences and anatomical regions.

\end{abstract}

\newpage

\section*{Introduction}
Pathology of the human spine is a major contributor to pain, disability, and reduced quality of life worldwide, placing a substantial and increasing burden on individuals and healthcare systems. 
In 2020, an estimated $619$ million people were affected by low back pain, which is the leading cause among all conditions for years lived with disability globally; this number is projected to exceed $800$ million by 2050~\cite{vos2020global,hartvigsen2018low,GlobalEpidemicLowBackPain2023}. 
In parallel, spinal cord injury disproportionately affects older adults and is associated with profound declines in functional independence and quality of life~\cite{lu2025global}. Given the high prevalence and severe consequences of spinal disorders, timely and accurate diagnosis is critical, as delays in identifying conditions such as spinal stenosis, disc herniation, or spinal cord compression may result in progressive neurological deficits, irreversible functional impairment, and poorer long-term clinical outcomes~\cite{forsth2016randomized,hartvigsen2018low,milligan2019degenerative,katz2022diagnosis}. 

Magnetic resonance imaging (MRI) is critical to diagnostic evaluation of the spine because of its soft-tissue contrast providing detailed visualization of: discs, spinal cord, nerve roots, and surrounding structures without ionizing radiation~\cite{hashemi2010mri}. Spine MRI interpretation, however, is complex and time-consuming, requiring expertise. This difficulty is compounded by the growing shortage of diagnostic radiologists, as imaging demand continues to outpace workforce capacity~\cite{afshari2025growing}. In this setting, where reports must integrate findings across diverse sequences and orientations, AI systems have the potential to help structure complex information, highlight salient abnormalities, and prompt efficient and consistent clinical reporting.

Despite these promises, three major challenges hinder the development of automated report generation systems for spine MRI. 
First, unlike many other medical imaging modalities~\cite{wu2025vision,bluethgen2025vision,xu2024whole}, MRI comprises many sequences, including T1- and T2-weighted imaging, that capture complementary yet distinct tissue information critical to distinguishing pathology~\cite{hashemi2010mri}. Modeling all sequences with a single shared model risks obscuring critical sequence-specific diagnostic patterns. 
Second, most of the conventional multi-modal pre-training frameworks~\cite{radford2021learning,zhai2023sigmoid} are designed to align more heterogeneous data structures, such as images and text. In contrast, different MRI sequences depict the same anatomy from the same patient, and are therefore highly similar in structure. We found that directly applying image–to-image alignment across MRI sequences resulted in model collapse and unstable optimization. 
Third, MRI report generation is inherently complex~\cite{monshi2020deep}, requiring integration of many heterogeneous subtasks across anatomical structures and imaging sequences, including condition diagnosis, lesion localization, and comparison with clinically similar cases while rigidly conforming to expected reporting patterns. Consequently, a single monolithic model is often insufficient to capture full spectrum of diagnostic requirements.

To address these challenges, we propose \ourAG, a multi-agent framework designed for spine MRI report generation comprising three stages: pre-training, continual training, and multi-agent report-generation. 
In the first stage, we pre-train two sequence-specific base models for T1-weighted and T2-weighted imaging (the most frequently used sequences in spine imaging) using large-scale clinical data comprising $32{,}047$ patients and $453{,}683$ MRI series, totaling $13{,}441{,}191$ slices. This design enables the learning of high-quality sequence-specific representations. 
In the second stage, to incorporate additional MRI sequences that are less prevalent and lack sufficient data for standalone pre-training, we introduce a novel continual training strategy that learns a lightweight synthesis module. This module derives embeddings for other sequences by leveraging the pre-trained T1 and T2 encoders.
Together, these two stages produce a unified multi-sequence foundation model, \ourFM, capable of generating high-quality robust embeddings for arbitrary MRI sequences. 
In the third stage, we instantiate a collection of $37$ task-specific agents from the shared foundation model \ourFM, including $17$ diagnostic agents for spinal condition prediction, $17$ agents for pathological slice identification and lesion localization, $2$ agents for retrieval of clinically similar patients, and a dedicated Medical Report Agent. 
The outputs of the $36$ upstream agents of \ourAG, including embeddings, diagnostic predictions, segmentation results, retrieved reference cases, and expert-designed prompting templates, are consolidated as structured tokens and provided as inputs to the Medical Report Agent.
Built on a LLaVA-style architecture, this agent integrates multimodal signals to generate the final radiology report in an end-to-end manner.

We demonstrate the effectiveness, versatility, generalizability, and interpretability of \ourAG by evaluating it on $50{,}705$ scans across $37$ tasks. 
The framework achieves state-of-the-art performance in identifying $17$ spinal conditions, demonstrating strong multi-sequence MRI understanding. Beyond predictive performance, \ourAG provides condition-specific attribution across MRI sequences, offering clinically meaningful explanations aligned with radiologists' diagnostic reasoning. To support interpretability, \ourAG localizes condition-relevant slices and delineates pathological regions, facilitating transparent decision support.
We also validate the multimodal representation capability of \ourFM through cross-modal image–text retrieval, achieving a $56.38\%$ improvement in Recall@5 on the UW Medical Center dataset compared with the second-best approach. Finally, by integrating all $37$ specialized agents built on top of \ourFM, \ourAG attains state-of-the-art performance in radiology report generation. Clinical evaluation by expert radiologists, based on corrected reports and RADPEER scoring~\cite{goldberg2017acr}, indicates that the generated reports achieve satisfactory quality for in-scope cases and can support clinical workflows when used with expert oversight. Here, clinical workflow refers to routine radiology report drafting with AI-assisted preliminary reports subject to radiologist review, and expert oversight denotes final verification and correction by trained radiologists prior to clinical use.

In summary, \ourAG leverages a unified multi-sequence MRI foundation model \ourFM, and a set of clinically grounded agents, to achieve state-of-the-art performance across $37$ diverse tasks, including the generation of high-quality spine MRI reports.

\begin{figure}[!htb]
    \centering
    \includegraphics[width=1\linewidth]{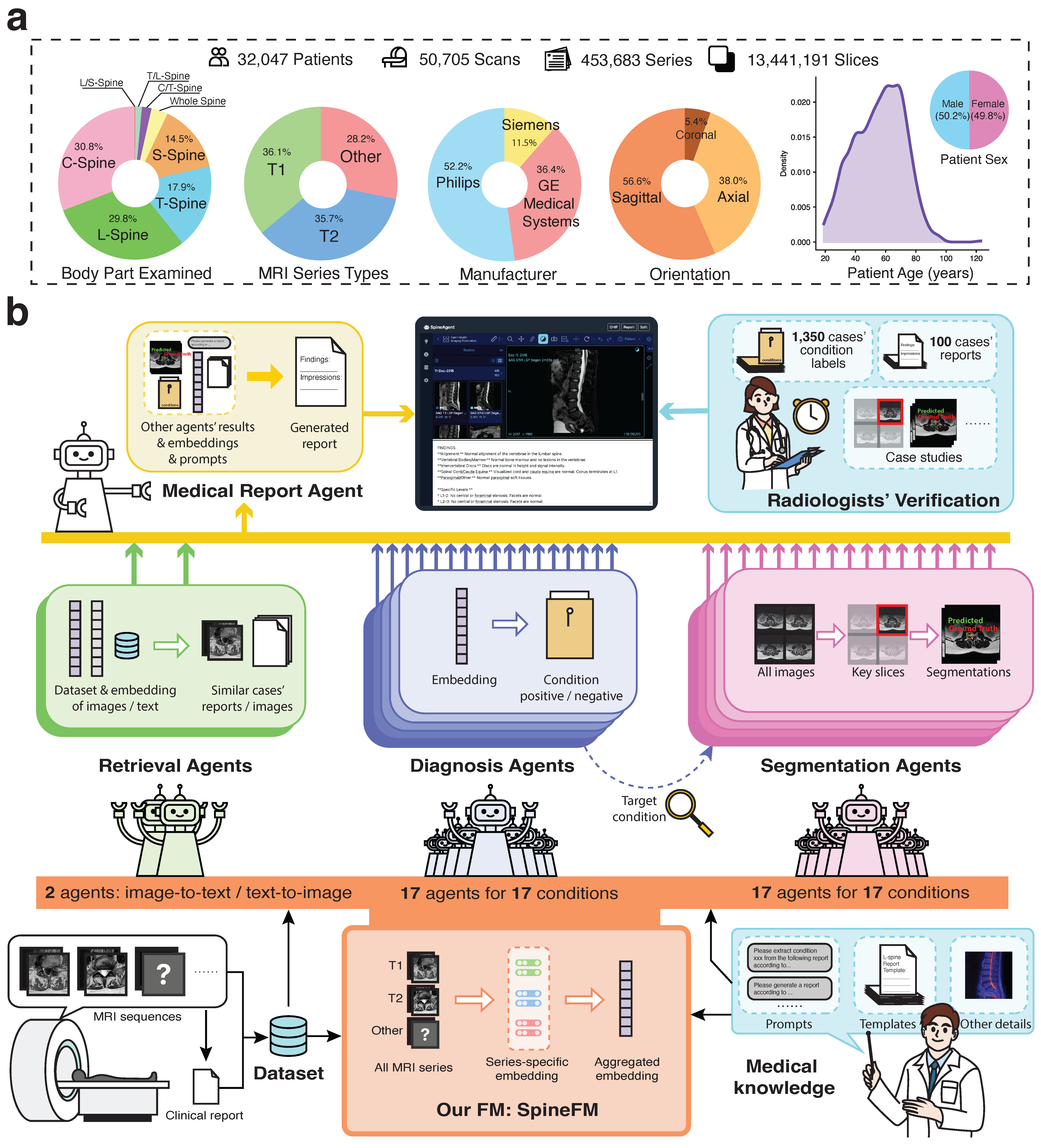}
    \caption{Overview of the {\ourAG} framework consisting of a multi-sequence MRI foundation model {\ourFM} and a multi-agent framework \ourAG for report generation. \textbf{a,} Our pre-training data collected at UW Medical Center contain $32{,}047$ patients and $453{,}683$ series, covering diverse body parts, manufacturers and patient age groups. 
    \textbf{b}, \ourFM is first trained as a multi-sequence foundation model capable of encoding heterogeneous MRI inputs. It then supports a set of specialized agents that generate complementary diagnostic signals, whose outputs are integrated to guide the Medical Report Agent for final report generation.}
    \label{fig:1}
\end{figure}

\section*{Results}

\subsection*{Overview of {\ourAG}}

\ourFM, the multi-sequence MRI foundation model underlying \ourAG, is pretrained on a large corpus of unlabeled MRI data from UW Medical Center (\textbf{Fig.~\ref{fig:1}a}), comprising $453{,}683$ MRI series from $32{,}047$ patients ($13{,}441{,}191$ slices). The dataset spans diverse anatomical regions, imaging orientations, and MRI series types, and includes scans acquired across multiple manufacturers and field strengths, capturing substantial real-world variation in image quality and acquisition protocols. This scale and diversity enable \ourFM to learn robust and transferable MRI representations. 
We evaluate \ourAG on large-scale benchmarks and through expert review: five radiologists independently revise generated reports and assign RADPEER scores~\cite{goldberg2017acr}, enabling structured assessment of clinical correctness and potential impact on patient management.

\ourAG introduces two key technical advances. First, while \ourAG supports a wide range of tasks (Fig.~\ref{fig:1}b), including condition prediction, segmentation, and retrieval, our primary focus is radiology report generation. We develop \ourAG as a multi-agent framework that integrates diagnostic predictions, spatial localization cues, and retrieved clinically similar cases within a unified reporting pipeline.
A Medical Report Agent based on a LLaVA-style architecture ingests these heterogeneous signals as structured token inputs, generating reports that are more accurate, anatomically consistent, and clinically grounded. Second, beyond the multi-agent design, \ourFM incorporates a sequence-aware continual training strategy to model the distinct diagnostic roles of different MRI sequences. In particular, \ourFM first trains two sequence-specific base encoders for T1- and T2-weighted imaging, and subsequently leverages continual learning to adaptively embed additional MRI sequences on top of the pretrained T1 and T2 encoders. This design captures sequence-dependent semantics while maintaining flexibility across diverse MRI protocols.

\subsection*{{\ourAG} accurately predicts spinal conditions}

\begin{figure}[!htb] 
    \centering
    \includegraphics[width=1\linewidth]{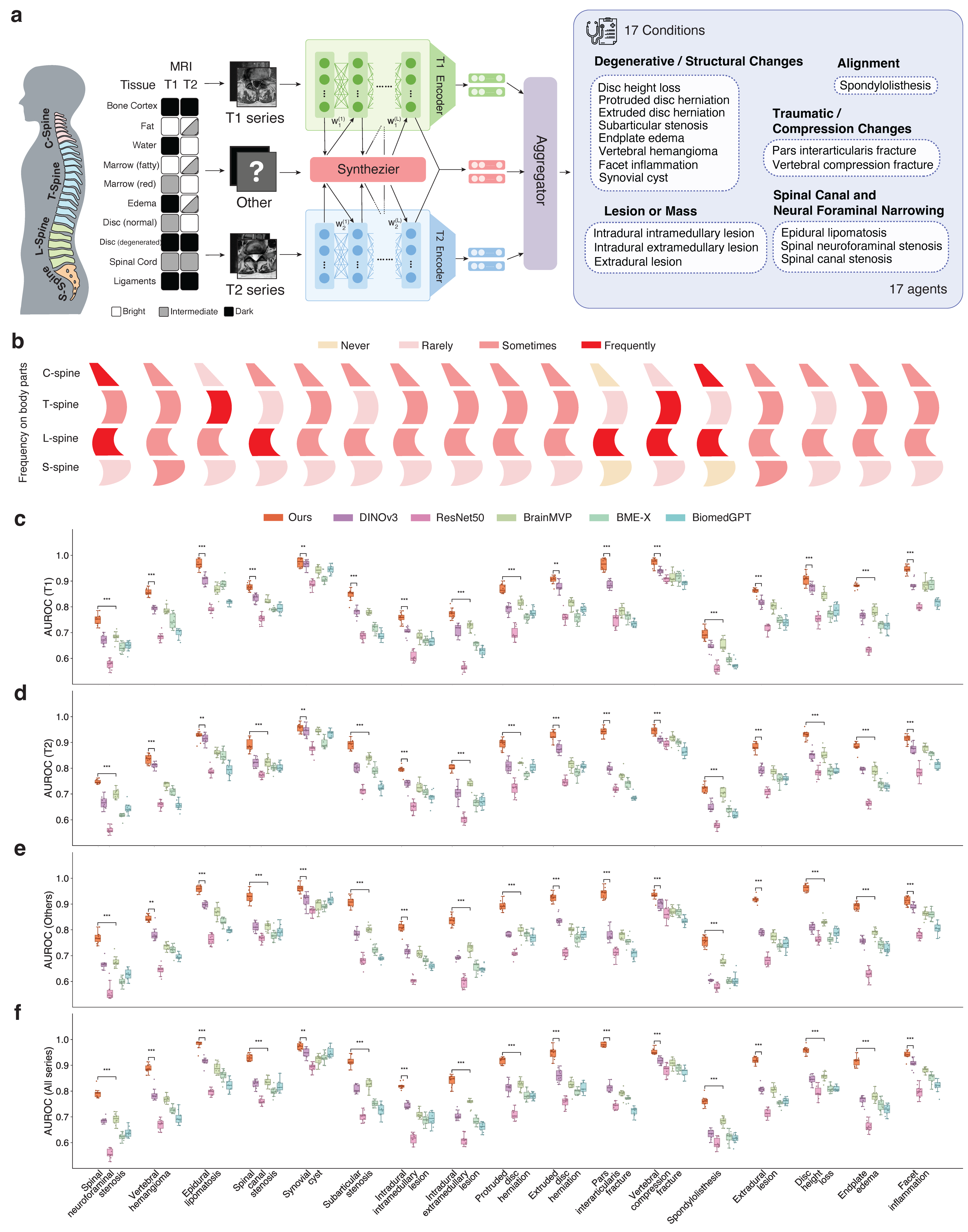}
    \caption{\textbf{Evaluation of \ourAG on spinal condition classification across $17$ conditions. \textbf{a,}} Overview of the multi-modal continual pretraining framework of \ourFM. 
    T1- and T2-weighted images are encoded using dedicated encoders, while other sequences are embedded via a synthesis module derived from T1 and T2 representations. 
    \textbf{b,} The prevalence of the $17$ conditions across four spinal regions, aligned with panels \textbf{c--f}. \textbf{c--f,} AUROC comparison of \ourAG Diagnosis Agents and baseline models trained on T1-only, T2-only, non-T1/T2, or all sequences. }
    \label{fig:2}   
\end{figure}

We first evaluate \ourAG's Diagnosis Agents on $1{,}350$ patients with radiologist-annotated labels for $17$ spinal conditions (\textbf{Fig.~\ref{fig:2}}), using our customized online system (\textbf{Supplementary Fig.~\ref{fig:supp_labelingsystem}}). These conditions span multiple clinically relevant categories (\textbf{Fig.~\ref{fig:2}a}), including degenerative or structural changes, alignment abnormalities, lesions or masses, traumatic or compression-related findings, and spinal canal or neural foraminal narrowing. This diverse set enables a comprehensive evaluation of \ourAG as a generalist framework for spinal MRI image analysis. We compare \ourAG against methods with general-domain vision encoders, such as DINOv3~\cite{simeoni2025dinov3} and ResNet50~\cite{he2016deep}, state-of-the-art MRI models, BrainMVP~\cite{rui2024brainmvp}, BME-X~\cite{sun2025foundation}, as well as the biomedical vision-language model BiomedGPT~\cite{zhang2024generalist,peng2025scaling}. Evaluation is conducted using a patient-level split with $10$-fold cross-validation.

\begin{figure}[h]
    \centering
    \includegraphics[width=1\linewidth]{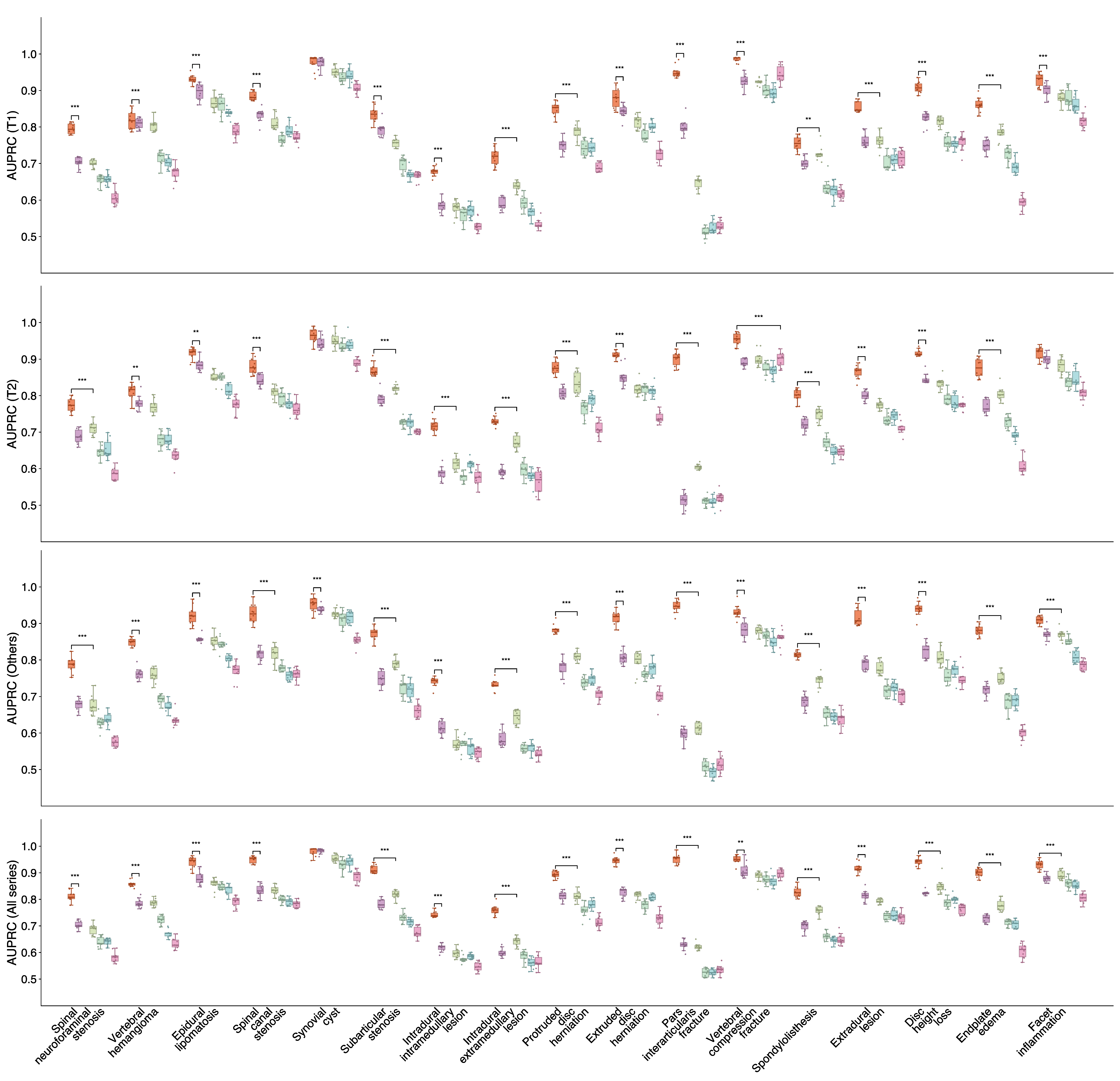}
    \suppfigurecaption{\textbf{AUPRC comparison} between \ourAG Diagnosis Agents and competing models across different settings, trained on only T1-weighted, only T2-weighted, non-T1/T2 (other) sequences, or all sequences combined. \ourAG consistently achieves superior performance across all settings. }
    \label{fig:supp_diagnosis_auprc}
\end{figure}

Across all conditions and metrics, \ourAG consistently outperforms competing methods by a substantial margin.
On T1- and T2-weighted sequences, for which \ourFM is explicitly pre-trained, it achieves average improvements of $8.79\%$ in AUROC (\textbf{Fig.~\ref{fig:2}b--e}) and $11.9\%$ in AUPRC (\textbf{Supplementary Fig.~\ref{fig:supp_diagnosis_auprc}}) over the strongest baseline. Importantly, for other MRI sequences, which constitute $28.2\%$ of the series and are not included in the initial pre-training stage, \ourAG's Diagnosis Agents still yields average gains of $11.1\%$ in AUROC and $13.7\%$ in AUPRC through continual training alone, underscoring the effectiveness and robustness of our proposed continual learning strategy in \ourFM.
Finally, when evaluated jointly across all sequences, \ourAG achieves improvements of $10.8\%$ in AUROC and $13.4\%$ in AUPRC, demonstrating its ability to integrate complementary information across sequences for accurate spinal condition modeling.

\begin{figure}[!htb] 
    \centering
    \includegraphics[width=1\linewidth]{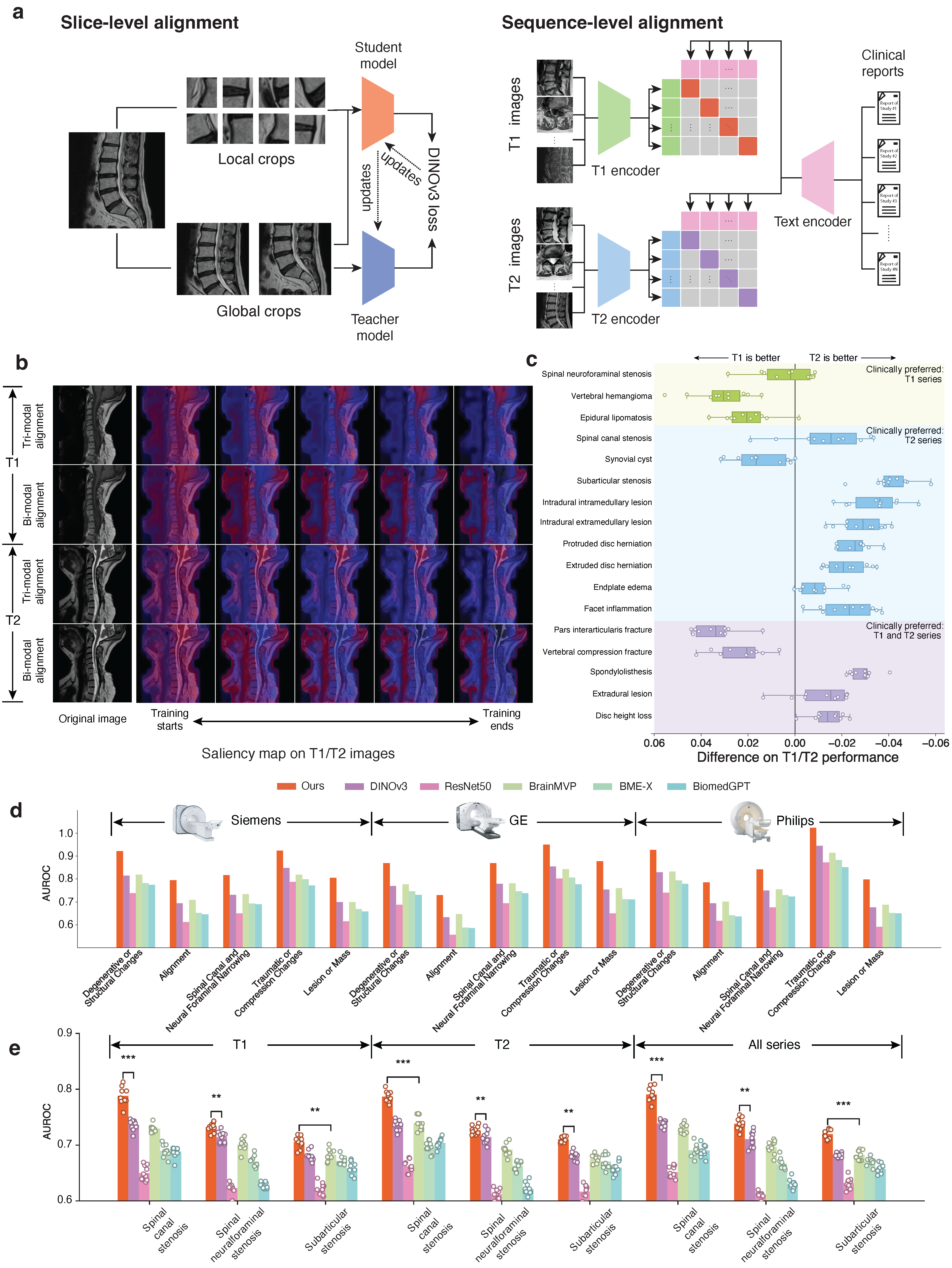}
    \caption{\textbf{\ourFM enhances generalizability via multi-modal continual training. }
    \textbf{a,} \ourFM integrates slice-level pretraining (DINOv3) and sequence-level continual training. The former captures multi-scale visual features, 
    while the latter aligns multi-modality features using clinical reports.  
    \textbf{b,} Saliency maps under different training strategies. Bi-modal emphasizes less informative structural features, whereas tri-modal highlights pathology-relevant regions, with divergence increasing over training.
    \textbf{c,} AUROC differences between T1- and T2-weighted models across conditions (T1 minus T2). 
    \textbf{d--e,} Cross-manufacturer and cross-cohort evaluation of \ourAG Diagnosis Agents, respectively.
}
    \label{fig:3}
\end{figure}

\subsection*{\ourFM capture sequence-specific patterns through multi-modal continual training}

\ourFM serves as the foundational backbone of \ourAG, providing shared representations that underpin all downstream agents.
In addition to modeling arbitrary MRI sequences, \ourFM leverages clinical reports to align representations derived from T1- and T2-weighted images. Unlike vision-language alignment in CT or X-ray imaging, which typically involves a single modality, multi-sequence MRI consists of acquisitions with distinct imaging characteristics. 
For example, T1-weighted images primarily delineate anatomical structure and tissue composition, whereas T2-weighted images emphasize pathological changes such as edema and inflammation.
These complementary yet diagnostically distinct properties motivate the use of separate base encoders for T1- and T2-weighted sequences.
This separation, however, raises a key challenge on how to align representations across sequences to enable unified, patient-level predictions that integrate information from all available MRI series (\textbf{Fig.~\ref{fig:3}a}).

\begin{figure}[h]
    \centering
    \includegraphics[width=1\linewidth]{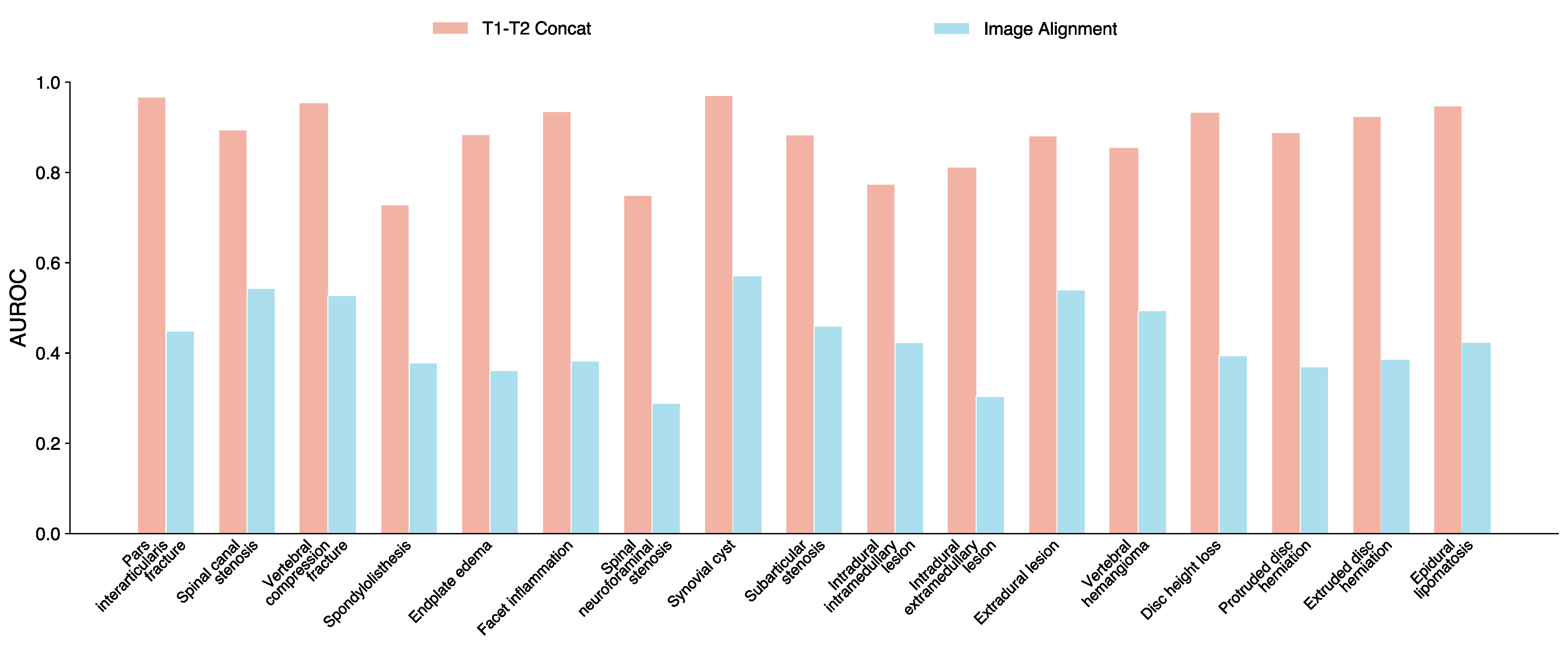}
    \suppfigurecaption{E\textbf{vidence that image-to-image bi-modal alignment is suboptimal.} Comparison of AUROC scores for combining \ourFM's T1- and T2-weighted representations via direct feature concatenation versus after a CLIP-style alignment between T1- and T2-weighted imaging modalities. Direct image-to-image alignment yields lower performance than simple concatenation, suggesting challenges in aligning heterogeneous MRI sequences.}
    \label{fig:supp_img2img_vs_concat}
\end{figure}

To address this problem, we evaluated multiple architectures for cross-sequence alignment. A tri-modal strategy that aligns T1 and T2 embeddings through clinical reports consistently outperformed a bi-modal approach that directly aligns T1 and T2 images, as commonly adopted in CLIP-style image–image alignment~\cite{radford2021learning}. Notably, direct image–image alignment underperformed even a simple concatenation of T1 and T2 embeddings (\textbf{Supplementary Fig.~\ref{fig:supp_img2img_vs_concat}}).
Saliency map analysis provides insight into this behavior, by comparing bi-modal (image–image) alignment and tri-modal (image–text–image) alignment strategies. 
As training progresses, the T1 and T2 encoders naturally diverge to reflect their distinct imaging characteristics. Bi-modal alignment encourages convergence toward visually consistent but diagnostically less informative features, such as global anatomical contours. In contrast, tri-modal alignment highlights disease-relevant regions and supports more effective cross-sequence integration (\textbf{Fig.~\ref{fig:3}b}).

The success of tri-modal alignment further motivates analysis of the relative contributions of different MRI sequences across spinal conditions. 
We quantify sequence-specific importance using relative AUROC differences between models trained exclusively on T1-weighted or T2-weighted series. 
To assess clinical plausibility of \ourFM's behavior, we examine whether these patterns are consistent with radiologists’ diagnostic preferences, given that T1- and T2-weighted images are relied upon to varying degrees depending on the condition. 
Across all $17$ spinal conditions, the observed performance differences closely match established clinical intuition (\textbf{Fig.~\ref{fig:3}c}), indicating that \ourFM captures sequence-specific representations aligned with expert diagnostic practice.

\subsection*{\ourAG demonstrates strong generalization across manufacturers and cohorts}

Next, we assessed the generalizability of \ourAG's Diagnosis Agents across heterogeneous scanner manufacturers and patient cohorts through cross-manufacturer and cross-cohort evaluations. In the cross-manufacturer setting, the model was trained exclusively on MRIs acquired from a single scanner manufacturer and evaluated on images from all manufacturers. We observed that \ourAG consistently achieved the best performance across five aggregated condition categories derived from all the $17$ spinal conditions (see definition in \textbf{Fig.~\ref{fig:2}a} and detailed performance of each condition listed in \textbf{Supplementary Fig.~\ref{fig:supp_crossmanufacturer}}), demonstrating strong robustness to scanner variability (\textbf{Fig.~\ref{fig:3}d}). 
The largest performance gains were observed in the ``Lesion or Mass'' category, consistent with the heterogeneous imaging appearances of these conditions across patients and acquisition platforms, which is more likely benefit from robust, scanner-invariant representations learned through large-scale pre-training.

We further evaluated cohort-level generalizability using the publicly available RSNA spine MRI dataset~\cite{richards2025rsna,lee2024performance,stephens2024rsna}. 
To simulate a cross-cohort scenario, the model was first fine-tuned using our UW Medicine dataset labels and then evaluated entirely on the RSNA cohort. \ourAG achieved the best performance across all three spinal conditions available in RSNA dataset, under T1-weighted, T2-weighted, and all-series settings, underscoring the effectiveness of the proposed pre-training framework. Consistent with trends observed in \textbf{Fig.~\ref{fig:2}e}, models trained using all available MRI series yielded the strongest overall performance. 

Together, these findings demonstrate that \ourAG maintains robust performance across both scanner manufacturers and patient cohorts, supporting its potential for deployment in diverse real-world clinical environments.

\begin{figure}[H]
    \centering
    \includegraphics[width=1\linewidth]{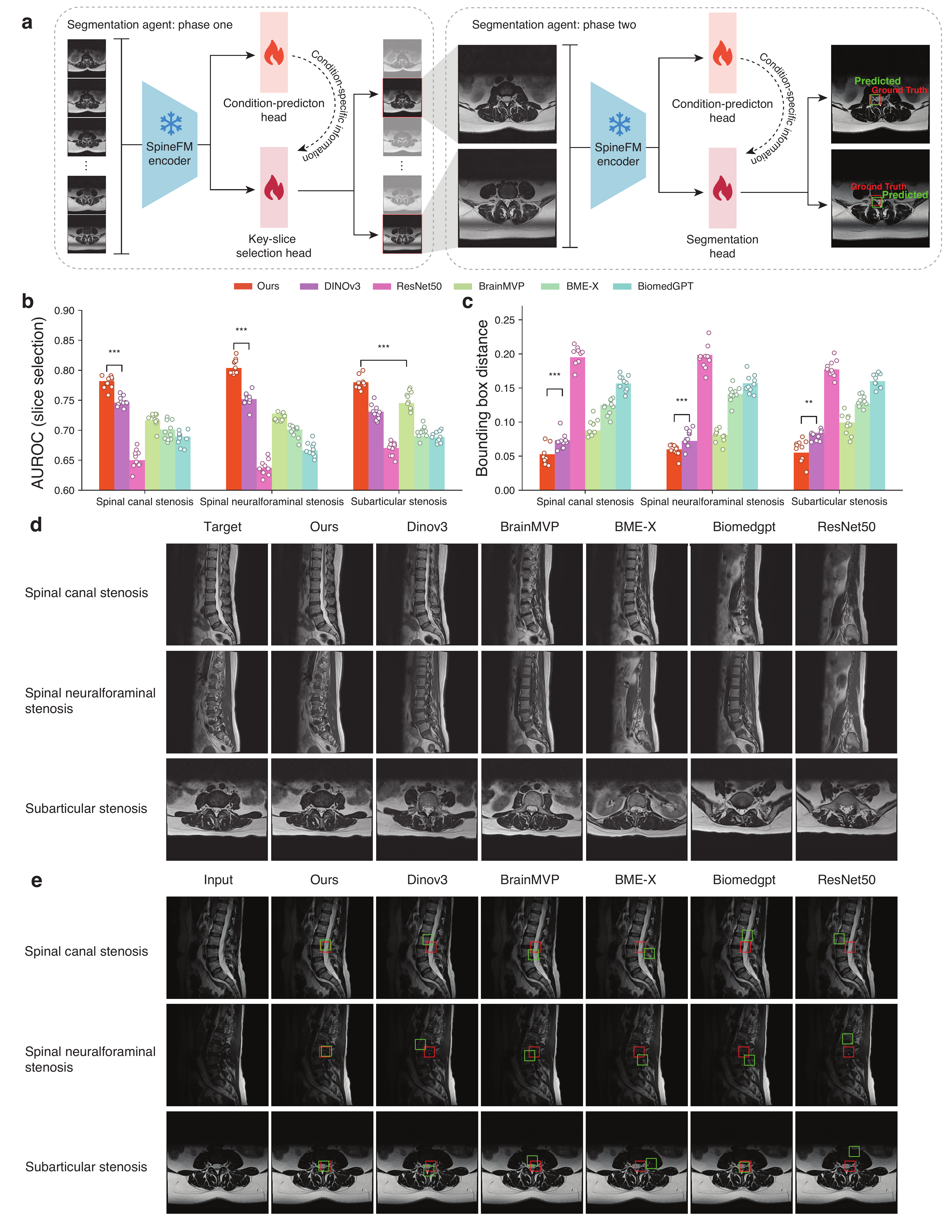}
    \caption{\textbf{Segmentation agents that localize pathological regions.} \textbf{a,} Segmentation agents' conditional-specific slice selection phase (left) and segmentation phase (right) find pathological regions with a 3D MRI image. Task heads differ by target condition (one of $17$). \textbf{b,} Performance on pathological slice selection in terms of AUROC. \textbf{c,} Performance on segmentation in terms of the distance between curated bounding box and predicted bounding box. \textbf{d,} Case studies showing the pathological slice selection by different methods. \textbf{e,} Case studies showing the segmentation performance by different methods. Red (/green) box indicates ground truth (/predicted) box.}
    \label{fig:4}
\end{figure}

\subsection*{\ourAG pinpoints pathological regions within 3D MRI series}

\ourFM's multi-modal continual learning framework further enables localization of pathological regions within three-dimensional MRI data, providing a critical component for report generation as implemented in the Segmentation Agents of \ourAG (\textbf{Fig.~\ref{fig:4}a}). 
Because MRI studies comprise multiple slices spanning broad anatomical coverage, we adopt a two-phase localization strategy in each Segmentation Agent. 
First, a slice-selection phase identifies slices that are most informative for the target condition. 
Second, a segmentation phase delineates the corresponding pathological regions within the selected slices using bounding boxes. 
This hierarchical design directs \ourAG's attention to clinically relevant locations across diverse MRI series and generates structured spatial cues for downstream report generation.

We evaluate slice selection and segmentation on the RSNA spine MRI benchmark, where \ourAG and all baselines are fine-tuned and assessed using the same data split. 
Slice selection aims to detect slices containing pathological patterns, whereas segmentation localizes the corresponding regions through bounding boxes. 
Across all three evaluated conditions, \ourAG consistently outperforms both general-domain vision foundation models and MRI-specific baselines, demonstrating strong capability in pathological slice identification and region delineation. 
Representative case studies further illustrate the advantages of \ourAG in slice selection (\textbf{Fig.~\ref{fig:4}d}) and segmentation (\textbf{Fig.~\ref{fig:4}e}), showing close correspondence between model predictions and clinically meaningful regions.

\subsection*{{\ourAG} enables cross-modal image retrieval}

Providing clinically similar reference cases may assist the Medical Report Agent of \ourAG in generating more accurate and contextually grounded radiology reports. To this end, we develop image-to-text and text-to-image Retrieval Agents that help us select relevant prior cases from the dataset. 

Given MRI inputs and a pool of candidate reports, including $4{,}215$ cases from the UW Medical Center dataset and $510$ cases from the RSNA dataset, the Retrieval Agent built on \ourFM achieves the strongest performance, substantially outperforming all baseline models on image-to-text retrieval tasks. 
This advantage is consistently observed on both the internal UW Mediccal Center cohort (\textbf{Fig.~\ref{fig:5}a}) and the external RSNA dataset (\textbf{Fig.~\ref{fig:5}b}), demonstrating the robust cross-dataset generalization of \ourAG.

Aside from case-level report retrieval, we examine slice-level image–text retrieval performance of the text-to-image Retrieval Agent. Given an input MRI slice, \ourAG retrieves the most relevant textual report, and we qualitatively compare the corresponding slices from the retrieved study at matched anatomical locations to assess visual and pathological consistency. Relative to all baselines, \ourAG more consistently retrieves reports associated with anatomically and pathologically concordant images, indicating superior image–text representation alignment (\textbf{Fig.~\ref{fig:5}c}).

We also analyze the contributions of different sequence-specific encoders in \ourAG's retrieval performance. As expected, the T1 encoder achieves the best results when retrieving T1-weighted images, whereas the T2 encoder performs best on T2-weighted series. Notably, for other sequences, the synthesizer module, which generates embeddings beyond T1 and T2, yields the strongest retrieval performance (\textbf{Fig.~\ref{fig:5}d}), further supporting its effectiveness in representing non-T1/T2 MRI sequences.

\subsection*{{\ourAG} for Automated Report Drafting Support}

\begin{figure}[H]
    \centering
    \includegraphics[width=1\linewidth]{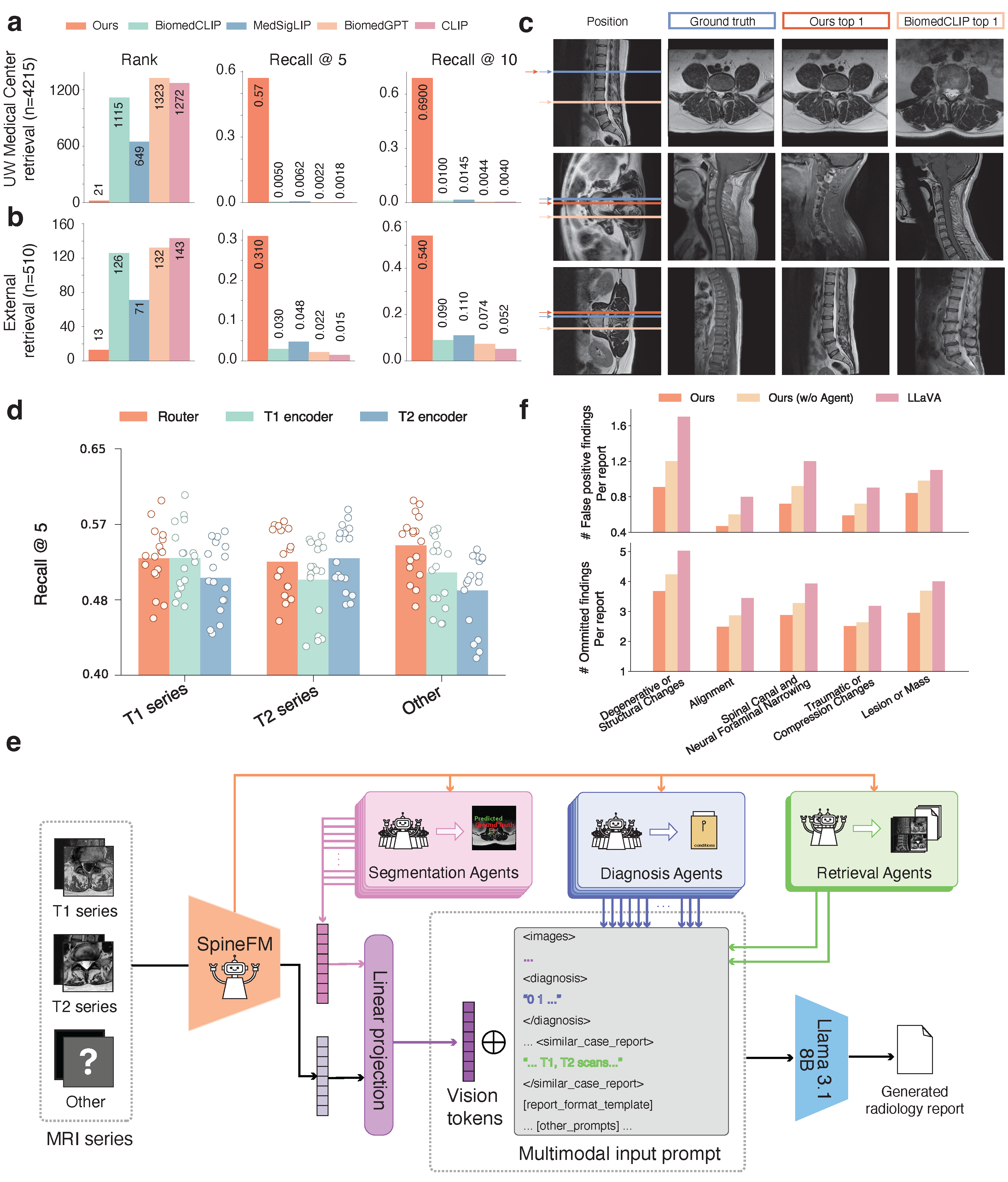}
    \caption{\textbf{Retrieval performance of {\ourFM} compared with baseline models.} \textbf{a,} Intra-dataset case retrieval on the internal UW Medicine cohort. \textbf{b,} Intra-dataset case retrieval on the external RSNA cohort. \textbf{c,} Cross-orientation image retrieval across different imaging planes. \textbf{d,} Recall @ $5$ performance stratified by MRI series type, comparing models that employ the synthesizer-based visual encoder with those that do not. \textbf{e,} Pipeline for generating radiology report drafts from MRI series. \textbf{f,} Quantitative comparison of report drafting performance between \ourAG and baseline methods.}
    \label{fig:5}
\end{figure}

\begin{figure}[!htb]
    \centering
    \includegraphics[width=1\linewidth]{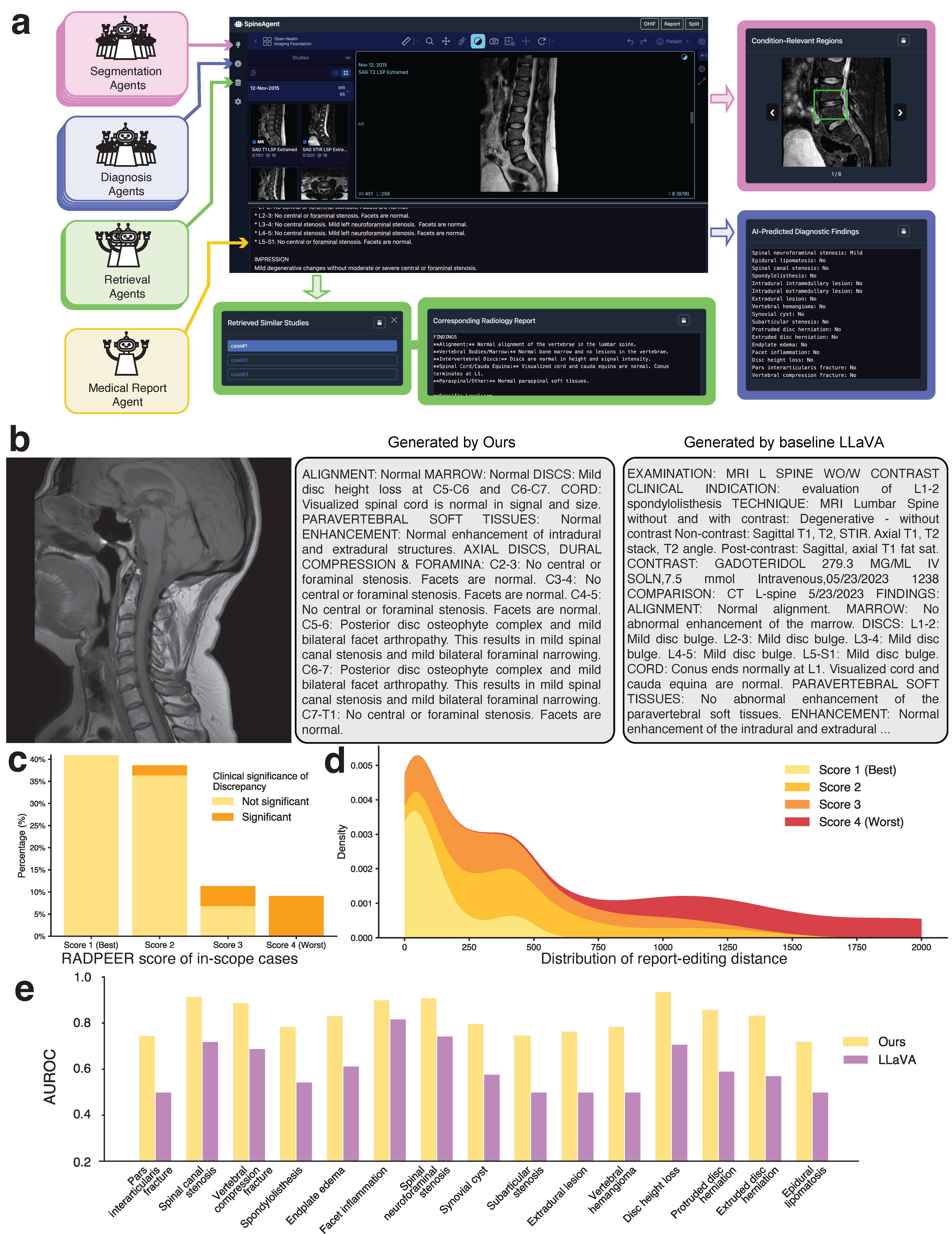}
    \caption{\textbf{Medical-report-drafting performance with \ourAG.} \textbf{a,} Screenshot of a demo case of our online tool using \ourAG to help with radiology-report generation. \textbf{b,} Representative examples of generated report drafts. \textbf{c,} Distribution of the RADPEER score of the in-scope cases. \textbf{d,} Distribution of the editing distances of the in-scope cases' reports. \textbf{e,} Comparison of \ourAG and a baseline LLaVA model, using human-corrected reports as the gold standard.}
    \label{fig:6}
\end{figure}

Instead of treating report generation as a black-box mapping from MRI series to text, we decompose the process into interpretable steps, including key-region segmentation, condition prediction, and retrieval of reports from clinically similar cases. Human expert knowledge is incorporated as structured text, and all components are converted into token sequences and combined into a multimodal prompt for a Llama-based model to draft radiology reports (\textbf{Fig.~\ref{fig:5}e}).
Compared with a baseline pretrained and adapted LLaVA model~\cite{liu2023visual} that directly generates radiology reports from images using text prompts, {\ourAG} produces substantially fewer false-positive findings and fewer omitted findings (\textbf{Fig.~\ref{fig:5}f}).
Ideally, by coordinating all agents within the \ourAG system, diverse diagnostic cues are jointly integrated to generate comprehensive radiology reports. The intermediate outputs produced by individual agents provide a degree of interpretability for human experts, making the overall framework more transparent and explainable than conventional black-box models, and thereby enhancing its practical utility in clinical settings (\textbf{Fig.~\ref{fig:6}a}).
An example case is shown in \textbf{Fig.~\ref{fig:6}b}. Although the imaged region clearly corresponds to the cervical spine (C-spine), the LLaVA baseline incorrectly identifies it as lumbar spine (L-spine) and generates less relevant conclusions, whereas {\ourAG} produces more accurate and anatomically consistent findings.

We also asked human experts for help in validating the quality of the generated reports.
Five radiologists, including a second-year resident, one fellow, and three attending radiologists with $9$, $15$, and $24$ years of clinical experience, respectively, independently conducted a structured manual review. Evaluations followed a standardized protocol implemented in a dedicated online platform built on the OHIF Viewer~\cite{ziegler2020open} and customized for \ourAG (\textbf{Supplementary Fig.~\ref{fig:supp_reviewingsystem}}). For each report, reviewers corrected inaccuracies, omissions and inappropriate phrasing, and assigned a RADPEER score~\cite{goldberg2017acr} to assess overall diagnostic quality and clinical acceptability, optionally providing case-specific comments when necessary.

Among the randomly selected MRI studies, a subset involves conditions or diagnostic details that were not represented in the training data. We refer to these as out-of-scope cases, as \ourAG was not explicitly trained to address such scenarios. In contrast, many cases fall within the predefined clinical targets for which structured supervision was provided during training. These in-scope cases correspond to the conditions and diagnostic categories that \ourAG is designed to model and support in routine practice.
Performance on out-of-scope cases reflects the natural boundary of the current training distribution. Addressing such cases would require additional curated annotations, expanded condition coverage, and potentially more intensive human-in-the-loop refinement. However, the primary objective of this study is to evaluate \ourAG within its intended clinical scope. In-scope cases represent the conditions for which sufficient supervision, validation, and expert input were incorporated during development, and therefore constitute the most appropriate basis for assessing clinical utility. Focusing on these cases allows a rigorous evaluation of the system's effectiveness in the settings for which it is explicitly designed, while acknowledging that expanding coverage to broader or rarer conditions remains an important direction for future work.

\textbf{Fig. \ref{fig:6}c} presents the distribution of RADPEER scores for in-scope cases. Most cases achieve favorable scores, indicating that the reports generated by \ourAG are generally of satisfactory quality when evaluated within the system's intended clinical scope.
\textbf{Fig. \ref{fig:6}d} shows the distribution of editing distances between \ourAG-generated reports and the corresponding human-corrected versions for the same in-scope cases. A clear trend emerges that cases with better RADPEER scores tend to require fewer edits. This relationship aligns with clinical intuition and suggests that higher-quality automated reports demand less post-editing. Together, these findings highlight the potential of \ourAG to assist radiologists by reducing reporting effort and accelerating workflow for in-scope cases. 
These findings are consistent with the design scope of our model, which achieves particularly strong performance on in-scope cases where structured supervision and expert guidance were incorporated during training. They also highlight opportunities for further refinement, including improving adherence to standardized reporting templates and enhancing the characterization of condition severity (\textbf{Supplementary Fig.~\ref{fig:supp_manual_feedback}}).
In \textbf{Fig.~\ref{fig:6}e}, we use human-corrected reports as the reference standard to evaluate how accurately reports generated by \ourAG capture case-specific conditions, in comparison with a baseline LLaVA model. Among the $17$ target conditions, randomly selected cases submitted for expert review did not include a sufficient number of intradural intramedullary or intradural extramedullary lesions to yield reliable AUROC estimates; these two conditions were therefore excluded, and evaluation was conducted on the remaining $15$. As shown, \ourAG consistently outperforms LLaVA across all evaluated conditions, indicating that its Medical Report Agent produces diagnostically more accurate reports.

\section*{Discussion}
We demonstrate that \ourAG consistently outperforms all competing baseline methods across the full set of $37$ agents on their respective tasks.
This performance advantage arises from several key design choices. 
First, unlike models pretrained on general-domain data, such as DINOv3~\cite{simeoni2025dinov3}, or on broadly defined biomedical corpora, such as BiomedGPT~\cite{zhang2024generalist,peng2025scaling}, \ourFM is trained on a large, clinically grounded corpus of spine MRI studies. \ourFM serves as the multi-sequence MRI foundation model underlying \ourAG. This domain-specific pretraining enables the model to learn representations tailored to spinal anatomy and pathology, rather than relying on generic visual or biomedical features.
Second, the scale of our pretraining dataset substantially exceeds that of publicly available spine MRI resources. In total, we leverage $13{,}441{,}191$ MRI slices acquired at UW Medical Center, providing extensive anatomical diversity, acquisition variability, and pathological coverage essential for robust representation learning.
Third, by adopting a foundation model paradigm, \ourAG effectively utilizes large volumes of unlabeled clinical data, alleviating reliance on scarce expert annotations. 
Finally, \ourAG integrates these representations within a coordinated multi-agent framework, allowing complementary agents to collaboratively address complex clinical tasks, most notably radiology report generation. Together, these elements underpin the strong and consistent performance of \ourAG across diverse spine MRI applications.

Despite these advances, several limitations warrant consideration and define directions for future work.
First, \ourAG is intended to assist rather than replace clinical experts, and all generated reports require final review by radiologists, consistent with current ethical standards and the need for human oversight in high-stakes medical decision-making. Enhancing interpretability, uncertainty quantification, and transparent reasoning over intermediate predictions may further strengthen clinical trust and enable more efficient human--AI collaboration.
Second, the current \ourAG framework is centered on MRI alone and does not incorporate complementary imaging modalities or broader clinical context. Integration of additional modalities, such as CT, as well as non-imaging data from electronic health records, may further improve diagnostic accuracy and report quality.
In addition, the present system operates on single examinations and does not explicitly model longitudinal imaging data, which could inform disease progression, treatment response, and temporal consistency. It is designed for diagnosis and reporting rather than preventive modeling. Extending this foundation model-based, multi-agent framework toward longitudinal risk stratification and early disease detection represents an important and promising avenue for future research.

Medical experts were closely involved throughout the design and development of \ourAG.
They contributed domain knowledge, including standard reporting templates used in routine radiology practice, detailed guidance on interpretation of spine MRI series, and the design of expert-verified prompts (see \textbf{Supplementary Prompts}). These prompts serve multiple purposes. For example, some are used to extract condition labels from structured human-written reports in our historical database, while others provide high-level clinical guidance to improve the quality and accuracy of report generation. Experts also participated in verification of diagnostic, segmentation, and condition-prediction outputs through manual case review. Most importantly, they conducted qualitative assessments of AI-generated radiology reports to check their clinical relevance, correctness, and usability.

\newpage
\section*{Methods}

\subsection*{Datasets}



The internal dataset used to pre-train \ourFM comprises $13{,}441{,}191$ MRI slices from $453{,}683$ series across $32{,}047$ patients, corresponding to $50{,}705$ imaging examinations acquired during routine clinical care at the University of Washington Medical Center.
The dataset spans multiple anatomical regions, MRI sequence types, imaging orientations, and scanner manufacturers, and includes a broad adult age range with a relatively balanced sex distribution. Data use was approved by the University of Washington Institutional Review Board (IRB), and all data were deidentified prior to analysis in accordance with the approved protocol.
In addition to DICOM images, deidentified radiology reports from the corresponding studies were incorporated under the same IRB approval.
Relevant clinical metadata were extracted from the electronic health record system in CSV format and processed to remove direct identifiers, with consistent deidentification applied to imaging metadata. 
The resulting large-scale, heterogeneous dataset was used exclusively for the development and training of \ourFM and \ourAG, and contains no protected health information.


For external evaluation, we used the RSNA Lumbar Degenerative Imaging Spine Classification (LumbarDISC) dataset~\cite{richards2025rsna}. This dataset comprises lumbar spine MRI studies from $2{,}697$ adult patients, totaling $8{,}593$ image series collected from eight institutions across six countries and five continents. 
Each study includes sagittal T1-weighted, sagittal T2-like (including T2, STIR, or Dixon~\cite{brandao2013comparing}), and axial T2-weighted series. Inclusion was restricted to outpatient degenerative disease cases, with exclusion of prior hardware, non-degenerative pathology, severe scoliosis, or substantial imaging artifacts. 
Expert annotators from RSNA, ASNR, and ASSR graded spinal canal, bilateral neural foraminal, and bilateral subarticular recess stenosis on a four-point scale (normal, mild, moderate, severe), with spatial localizers marking the corresponding spinal levels. Annotations were curated through standardized training, role-specific assignment, and consensus review for the test sets. These labels were subsequently collapsed into a three-class scheme by merging normal and mild grades. The dataset is partitioned into training ($1{,}981$ studies), public test ($272$), and private test ($444$) subsets, with balanced distributions across age, sex, site, and disease severity, and targeted enrichment of under-represented high-grade stenosis at specific levels. MRI data are released in DICOM format with accompanying CSV annotation files and are publicly available through Kaggle and RSNA MIRA. \footnote{Available at \url{https://www.kaggle.com/competitions/rsna-2024-lumbarspine-degenerative-classification}.}

\subsection*{Details of \ourAG}
\ourAG is a general-purpose, multi-agent system for spine MRI analysis, designed to support the diagnosis of a broad spectrum of spine-related conditions across heterogeneous MRI series. All agents in \ourAG are built upon a foundation model, \ourFM.

\ourFM is pretrained on a large corpus of unlabeled MRI data (\textbf{Fig.~\ref{fig:1}a}) from the University of Washington Medical Center. The dataset encompasses multiple anatomical regions, imaging orientations, sequence types, and scanner manufacturers, and spans a broad patient age range with a relatively balanced sex distribution.
This extensive diversity underpins the robustness and generalizability of the pretrained model \ourFM.

In contrast to conventional LLM-based agent systems that rely on textual reasoning within the language model to determine module selection~\cite{chowa2026language,shangagentsquare}, \ourFM adopts a routing-based design~\cite{fedus2022switch,rosenbaum2018routing} within the visual encoder. The resulting visual representations are shared across task-specific agents of \ourAG, whose outputs collectively inform a downstream Medical Report Agent. By integrating visual features and structured semantic signals from all agents, the report agent generates a draft radiology report to support spine clinicians in delivering accurate and efficient diagnoses.

\subsection*{T1 and T2 encoder pre-training}

At the pre-training stage, we first optimize the vision encoder components of \ourFM, namely the T1- and T2-specific encoders, on their corresponding MRI series using the DINOv3 framework~\cite{simeoni2025dinov3}. The T2-specific encoder processes both T2-weighted images and related sequences such as STIR and Dixon~\cite{brandao2013comparing}. We subsequently align visual and textual representations through a CLIP-based contrastive framework~\cite{radford2021learning}. Building upon these encoders, we train a lightweight router called synthesizer, that enables \ourFM to accommodate heterogeneous MRI series beyond canonical T1 and T2 inputs, followed by another phase of image–text alignment. Finally, visual representations and textual context are integrated within a LLaVA-based framework~\cite{liu2023visual} to support multimodal reasoning.

Both the T1 and the T2 vision encoders in \ourFM are based on the Vision Transformer (ViT) architecture~\cite{dosovitskiyimage} and are pre-trained using DINOv3 framework~\cite{simeoni2025dinov3}. The pre-training phase proceeds in two stages. In the first stage, the T1 and T2 encoders are independently trained on $4{,}501{,}930$ T1-weighted slices and $4{,}519{,}538$ T2-weighted slices (including STIR and Dixon), respectively. 
We use a per-GPU batch size of $100$ and train each model on four A100-80GB GPUs for $150$ epochs, corresponding to $187{,}500$ iterations. Following the default DINOv3 configuration, we adopt the standard DINO head, the KoLeo loss weight, and the iBOT loss:
\begin{equation}
    \mathcal{L}_\mathrm{\ourFM} = \mathcal{L}_\mathrm{DINO} + \mathcal{L}_\mathrm{iBOT} + \lambda \times \mathcal{L}_\mathrm{Koleo}
\end{equation}
We additionally introduce four register tokens into the input sequence during pre-training to enhance \ourFM's capacity for dense feature representation~\cite{darcet2023vision}. In the second stage, we apply gram anchoring by incorporating an additional gram anchoring loss, $\mathcal{L}{\mathrm{Gram}}$, into the pre-trained objective for a further $30$ epochs. 
All our training objectives, including the initial objective $\mathcal{L}\mathrm{\ourFM}$ and the gram anchoring loss $\mathcal{L}_{\mathrm{Gram}}$, are adapted from the DINOv3 framework~\cite{simeoni2025dinov3}.

\subsubsection*{Text-image alignment}

We performed CLIP training~\cite{radford2021learning} to enhance the multimodal representation capability of \ourFM. 
The pre-trained T1 and T2 encoders of \ourFM were used as the vision encoders, and BiomedBERT~\cite{chakraborty2020biomedbert} served as the language encoder. Contrastive learning was applied to align image and text embeddings in a shared latent space, encouraging paired representations to be close while separating mismatched pairs. 
The contrastive objective is defined as:
\begin{equation}\label{eq:clip}                           
\mathcal{L}_{\text{CLIP}} =
\frac{1}{2N} \left(
-\sum_{k=1}^{N}                                                                                                                     
\log \frac{\exp\left(\cos(\mathbf{f}_k, \mathbf{g}_k)/\tau\right)}
{\sum_{j=1}^{N} \exp\left(\cos(\mathbf{f}_k, \mathbf{g}_j)/\tau\right)}                                                                         
-\sum_{k=1}^{N}                               
\log \frac{\exp\left(\cos(\mathbf{f}_k, \mathbf{g}_k)/\tau\right)}                                                                              
{\sum_{j=1}^{N} \exp\left(\cos(\mathbf{f}_j, \mathbf{g}_k)/\tau\right)}
\right) \,,                                                                                                                                     
\end{equation}   
where $\cos(\cdot, \cdot)$ denotes the cosine similarity, $\mathbf{f}_k \in \mathbb{R}^d$ and $\mathbf{g}_k \in \mathbb{R}^d$ represent the image   
  and text embeddings for the $k$-th case, respectively, and $\tau > 0$ is an adjustable temperature parameter controlling the sharpness of similarity scaling. Lower values of $\tau$ increase sensitivity to subtle similarity differences and promote tighter alignment between matched image–text pairs.

The CLIP model was trained on four NVIDIA A100 GPUs (80GB), each with a local batch size of $4$. Gradient accumulation over $32$ steps resulted in an effective global batch size of $512$. Training was conducted for five epochs.

\subsubsection*{Continual training of the synthesizer for other series}


Given the substantial differences between T1- and T2-weighted MRI series, we train separate vision encoders tailored to their respective data distributions. For MRI series that cannot be reliably categorized as T1 or T2, we introduce a router-based fusion mechanism called synthesizer to produce series-agnostic embeddings. At each encoder layer, a Synthesizer module computes fusion weights that determine how intermediate representations from the T1 and T2 encoders are combined through a weighted sum. The fused representation is then propagated to the subsequent layer, enabling routing decisions to evolve hierarchically with network depth. This layer-wise fusion strategy allows dynamic adaptation to arbitrary MRI series while preserving sequence-specific inductive biases. The resulting series-level embeddings can be further aggregated across multiple series to form a unified case-level representation for downstream tasks.

During synthesizer training, we first optimize the visual encoder of \ourFM with the synthesizer module enabled using the full pre-training dataset from UW Medical Center, without restricting inputs to specific MRI sequence types. This stage promotes the learning of series-agnostic visual representations. 
To further align visual features with clinical semantics across heterogeneous modalities, we subsequently incorporate case-level radiology reports and continue training within the CLIP-based image–text contrastive framework (Equation~\ref{eq:clip}) once again, this time with the synthesizer remaining active throughout.

\subsubsection*{Multimodal generation of MRI report via LLaVA-style architecture}


To enhance the multimodal generation capability of \ourFM, we adopt the LLaVA framework~\cite{liu2023visual}, which integrates a vision encoder, a projector module, and a large language model (LLM) decoder. In our implementation, \ourFM serves as the vision encoder, and LLaMA-3.1-8B~\cite{grattafiori2024llama} functions as the language decoder.

Each patient study includes T1- and T2-weighted MRI sequences with variable numbers of slices. We first extract slice-level visual embeddings using the corresponding sequence-specific encoders. Embeddings from T1 and T2 sequences are concatenated and passed through an attention pooling–based projector, which aggregates variable-length slice-embeddings and maps them into a fixed number of image tokens. These image tokens are concatenated with text tokens derived from user instructions and jointly provided to the LLaMA decoder for multimodal generation.

The multimodal model is trained using a visual question-answering (VQA) dataset constructed from our radiology reports, comprising four task types: long-answer questions, short-answer questions, multiple-choice questions, and report generation.

Following the standard LLaVA training strategy \cite{liu2023visual}, we adopt a two-stage training protocol:
\begin{itemize}
    \item \textbf{Stage 1: Projector warm-up.}
    We freeze the vision encoder (\ourFM) and the LLaMA-3.1-8B language decoder, and fine-tune only the attention-pooling projector for two epochs with a learning rate of $1 \times 10^{-3}$. The global batch size is $32$, with gradient accumulation yielding an effective batch size of $128$.
    \item \textbf{Stage 2: Language decoder adaptation.}
    We then fine-tune the language decoder for four epochs using Low-Rank Adaptation (LoRA)~\cite{hu2022lora}, while keeping the vision encoder frozen. Specifically, LoRA adapters are injected into the LLaMA-3.1-8B~\cite{grattafiori2024llama} decoder to enable efficient parameter adaptation without updating the full model weights. The learning rate is set to $2 \times 10^{-5}$, with the same global and effective batch sizes as in Stage 1. This stage enables the language model to better align with the visual tokens produced by the projector and to perform multi-modal reasoning and report generation effectively.
\end{itemize}

\subsection*{Details of spinal condition prediction}


We focus on $17$ distinct clinically relevant spine condition defined and supervised by medical experts. For each condition, we construct a dedicated diagnostic agent to enable condition-specific learning and evaluation.

\begin{figure}[h]
    \centering
    \includegraphics[width=1\linewidth]{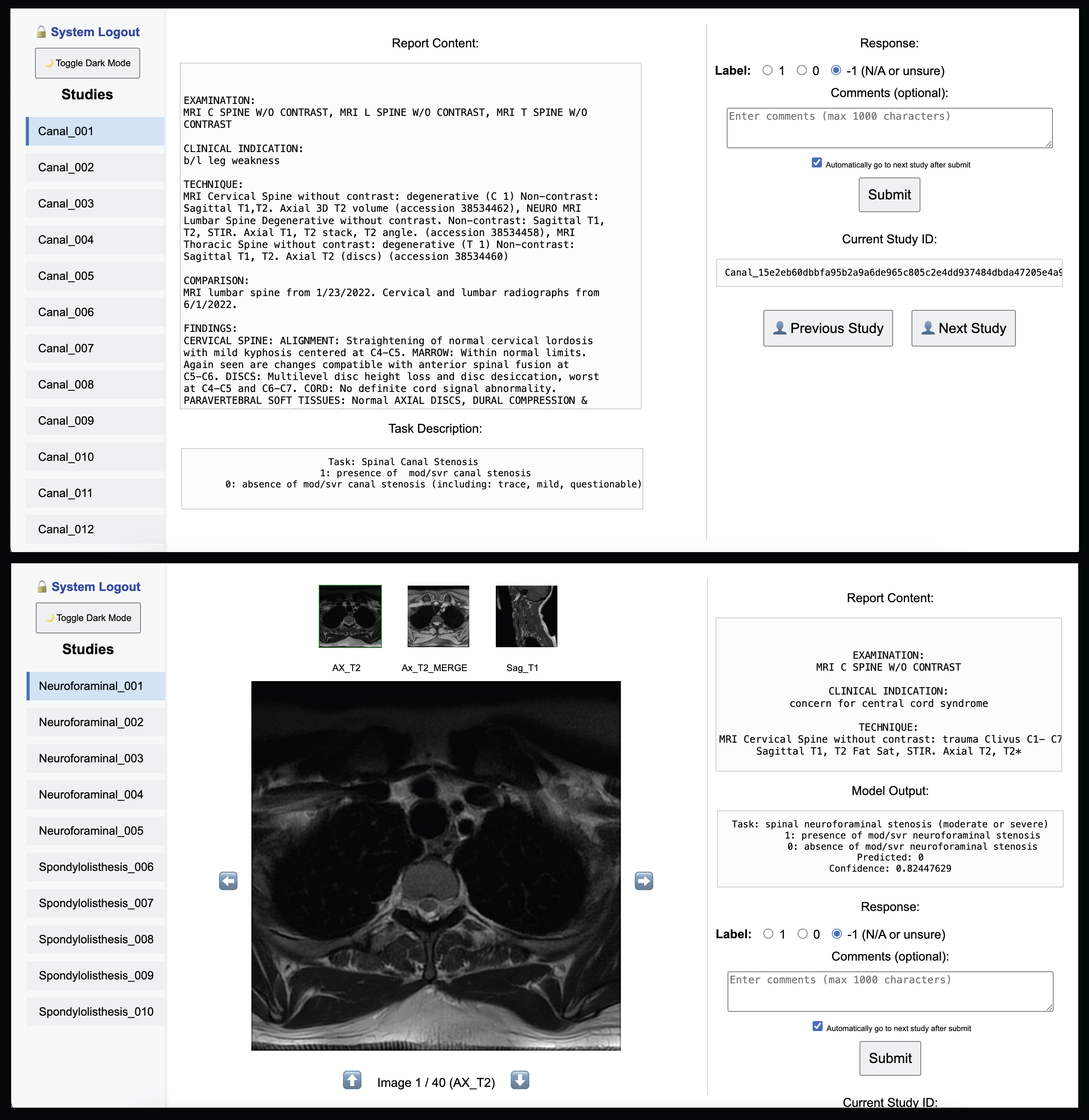}
    \captionsetup{labelformat=empty}
    \suppfigurecaption{\textbf{Screenshots of the manual annotation interface} used by radiologists to review and label selected studies from the UW Medical Center dataset. These annotations were used for reliable evaluation of \ourAG and baseline methods.}
    \label{fig:supp_labelingsystem}
\end{figure}

Condition-specific labels are derived from the radiology reports associated with each MRI study. For every target condition, domain experts design structured prompts to guide automated label extraction, producing ternary annotations that indicate presence ($1$), absence ($0$), or ambiguity ($-1$). Large language models, namely Llama-3.3-70B~\cite{grattafiori2024llama} and Phi-4-14B~\cite{abdin2024phi}, are used to apply these prompts at scale to the report corpus. 
To ensure clinical validity, a subset of $1{,}350$ automatically labeled cases was independently reviewed at the case level using an online system we specifically built for this project (\textbf{Supplementary Fig.~\ref{fig:supp_labelingsystem}}), by three radiologists at different stages of training, including a medical student, a second-year radiology resident, and a radiology fellow. All annotations were performed under the supervision of an attending radiologist with nine years of independent clinical practice. This review confirmed that the extracted labels were of sufficient accuracy and reliability for downstream model training and evaluation.

Condition prediction is formulated as a binary classification task, excluding ambiguous cases (labeled $-1$). We adopt a linear probing strategy, in which the parameters of the vision encoder backbone are frozen and a condition-specific linear classifier is trained on the corresponding case-level embeddings. Linear probing is widely used for evaluating foundation models, as it isolates the representational quality of learned embeddings while enabling efficient training and fair comparison across methods~\cite{chen2020generative}.

Because the condition labels are highly imbalanced, with substantially more negative ($0$) than positive ($1$) cases for most conditions, we report the area under the receiver operating characteristic curve (AUROC) as the primary evaluation metric for \ourAG and all baselines. AUROC provides a robust and threshold-independent measure of classifier performance under class imbalance~\cite{mcdermott2024closer}.

\subsection*{Details of pathological region identification}


We seek to localize anatomical regions associated with spine-related conditions in order to provide informative visual cues that guide radiology report generation.

Localization is implemented within the Segmentation Agents of \ourAG through a two-stage pipeline comprising slice selection followed by region segmentation. Reliable training and evaluation require MRI series with accurate annotations of both key slices and condition-specific regions. We therefore adopt the RSNA spine MRI benchmark, which provides expert-annotated key slices and bounding-box labels, enabling standardized supervision and quantitative comparison.

For slice selection, each MRI slice is treated as an independent sample and the task is formulated as a binary classification problem, where a label of $1$ indicates that the slice is condition-relevant and a label of $0$ indicates that it is not. The key-slice-selection head then predicts this binary label using condition-specific task heads attached to the corresponding visual encoder.

For region localization, a segmentation head regresses the spatial extent of the condition-associated region within each selected slice by predicting normalized bounding-box coordinates. This task is formulated as a regression problem over box parameters.

Together, this two-stage framework enables \ourAG to robustly identify informative slices and accurately localize condition-relevant regions across heterogeneous MRI series. 
There are $17$ Segmentation Agent variants instantiated, one per condition, each optimized for condition-specific characteristics. 

During evaluation, segmentation performance is assessed using a normalized bounding-box localization error, defined as the relative offset between predicted and ground-truth box coordinates, normalized by slice dimensions. This metric accounts for scale variation across images and provides a consistent measure of localization precision.

\subsection*{Details of cross-modal retrieval}

Retrieval is performed in a zero-shot manner, without any task-specific fine-tuning on the evaluated datasets. All retrieval experiments are explicitly conducted by the Retrieval Agents within \ourAG, which directly utilize the pretrained embeddings generated by \ourFM. This design enables assessment of the intrinsic semantic alignment learned during pretraining, independent of downstream adaptation.

For image-to-text retrieval, \ourAG's image-to-text Retrieval Agent encodes each MRI study into a case-level embedding and compares it against a candidate set of deidentified radiology report embeddings. Similarity scores are computed within the shared embedding space, and candidate reports are ranked accordingly. This setup reflects a clinically relevant use case scenario in which the system retrieves prior reports with similar imaging characteristics to support report drafting.

We further evaluate cross-orientation slice retrieval in a text-to-image setting using \ourAG's text-to-image Retrieval Agent. Given a textual query derived from a reference slice and its spatial context, the agent retrieves anatomically corresponding MRI slices acquired in different imaging planes from a candidate pool of studies. This evaluation probes the ability of the Retrieval Agent to align textual descriptions with anatomically consistent visual content across orientations and series types.

Retrieval performance is quantified using standard rank-based and recall-based metrics. We report the rank position of the correct match among all candidates and Recall@$k$, defined as the proportion of queries for which the correct match appears within the top-$k$ results. These metrics capture both fine-grained ranking quality and practical retrieval effectiveness. For stratified analyses across MRI series types, model variants with and without the synthesizer module are compared to evaluate its contribution to retrieval robustness under heterogeneous input conditions.

\subsection*{Details of MRI report generation}


Radiology report drafting in \ourAG is conducted explicitly by the Medical Report Agent and implemented as a coordinated, multi-stage multi-agent framework, rather than as a direct image-to-text mapping. This design is intended to reflect elements of clinical reasoning while improving interpretability, anatomical consistency, and diagnostic reliability in report generation.

For a given MRI study, all available series are first encoded into a shared representation using the pretrained visual backbone of \ourFM, enabling consistent processing of heterogeneous MRI inputs. These representations are subsequently processed by multiple specialized agents that extract complementary diagnostic signals, including condition-level predictions, spatial localization cues, and contextual information from clinically similar prior cases. Collectively, these agents provide structured and clinically meaningful evidence capturing both global and localized aspects of the study.

The aggregated outputs are integrated with expert-designed prompting templates and transformed into a multimodal input that conditions a large language model (Llama-3.1-8B~\cite{grattafiori2024llama}) to generate a draft radiology report. By grounding text generation in explicit diagnostic predictions, spatial context, and retrieved reference cases, the system promotes conservative synthesis of findings and anatomical coherence. The generated reports are intended to assist clinical interpretation rather than replace expert judgment, and randomly selected outputs are subsequently reviewed and verified by radiologists.


Report-drafting performance is evaluated using both automated and human-centered protocols. First, we employ a large language model–based evaluation framework to quantify factual errors, including false-positive findings, defined as hallucinated conditions not supported by imaging evidence, and omitted findings, defined as clinically relevant conditions present but not described. The evaluator compares generated reports against reference information derived from expert-validated condition labels and localization outputs.

\begin{figure}[h] 
    \centering
    \includegraphics[width=1\linewidth]{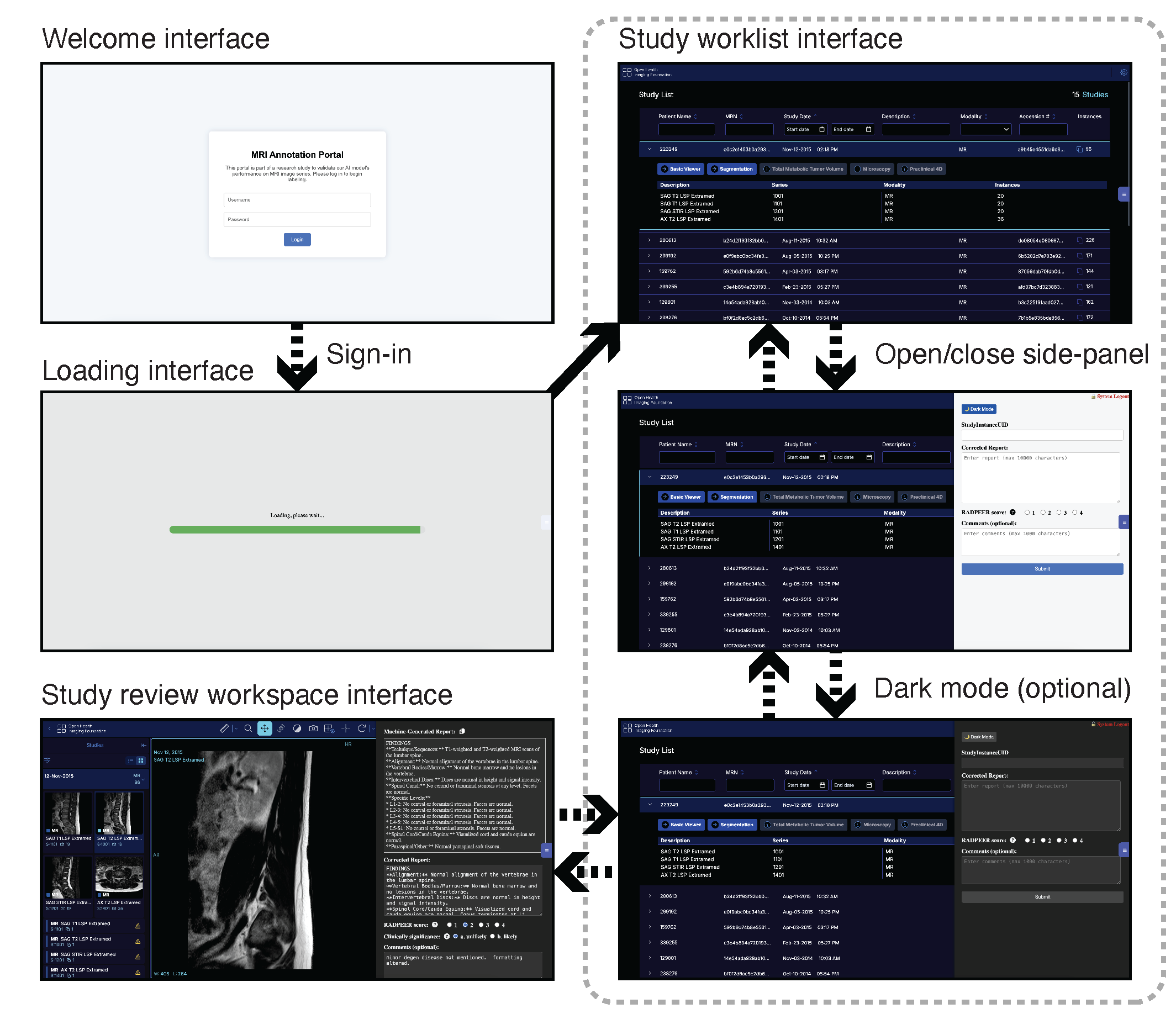}
    \captionsetup{labelformat=empty}
    \suppfigurecaption{\textbf{Screenshots of the report-review interface} used by radiologists to evaluate generated reports, provide corrected versions, assign RADPEER scores, and optionally record case-specific comments.}
    \label{fig:supp_reviewingsystem}
\end{figure}

\begin{figure}[h]
    \centering
    \includegraphics[width=1\linewidth]{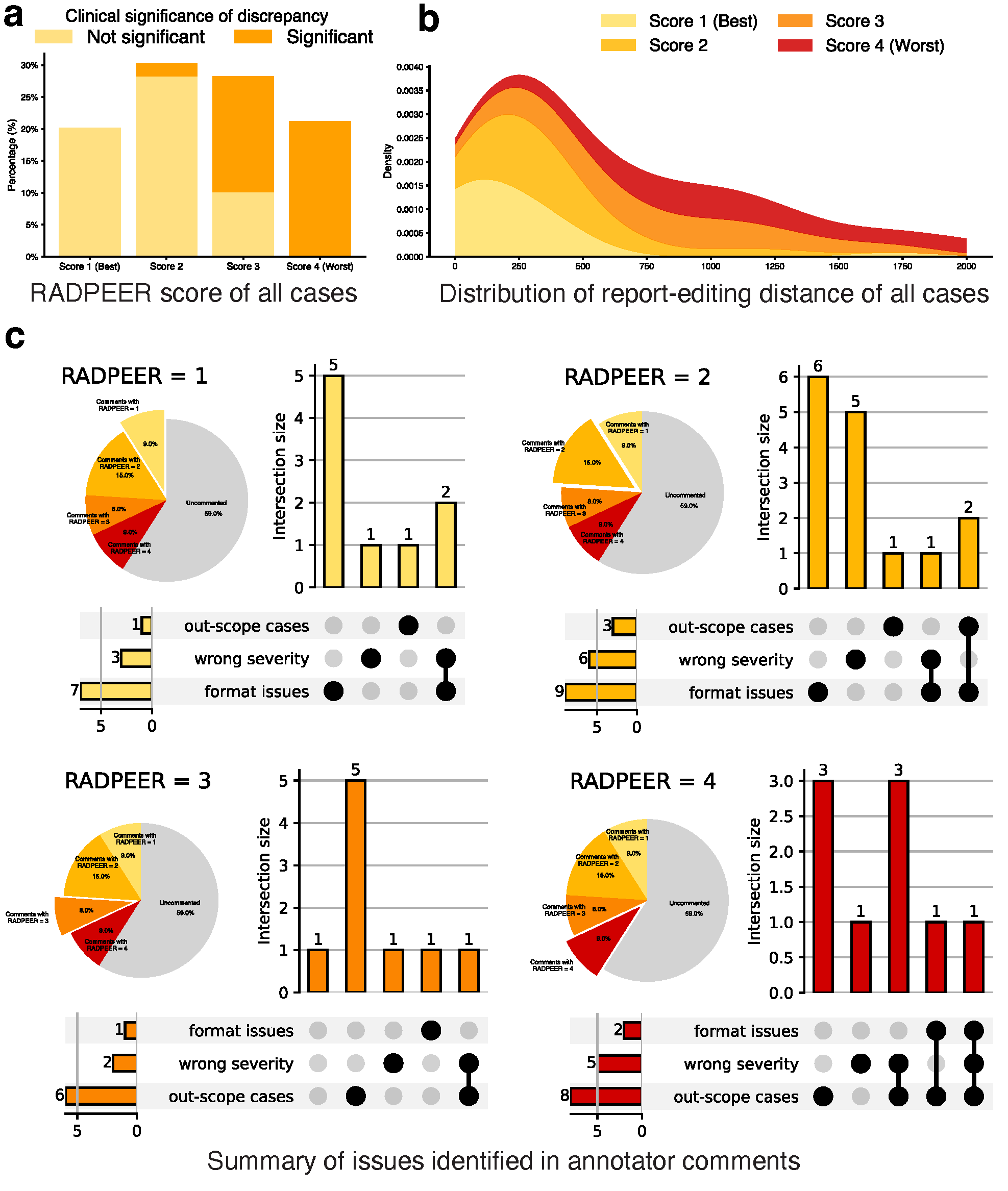}
    \captionsetup{labelformat=empty}
    \suppfigurecaption{\textbf{Summary of radiologists feedback across randomly selected $100$ cases}, including both in-scope and out-of-scope cases. \textbf{a,} RADPEER score distribution. \textbf{b,} Distribution of editing distances between generated reports and radiologist-corrected versions. \textbf{c,} Summary of issues identified in radiologist comments during case review and annotation, reflecting some of the limitations of the current version of \ourAG. Limited label coverage (that is, out-of-scope cases) is the primary contributor to worse RADPEER scores (like $4$ or $3$).}
    \label{fig:supp_manual_feedback}
\end{figure}

Second, we conduct a manual review by five radiologists, including a second-year resident, a radiology fellow, and three attending radiologists with $9$, $15$, and $24$ years of clinical experience, respectively. All reviewers follow a standardized protocol using a dedicated online evaluation system built upon the OHIF viewer~\cite{ziegler2020open} and specifically developed for reviewing \ourAG (\textbf{Supplementary Fig.~\ref{fig:supp_reviewingsystem}}). For each report, radiologists correct inaccuracies, omissions, and inappropriate phrasing, and assign a RADPEER score~\cite{goldberg2017acr} to assess overall diagnostic quality and clinical acceptability. This combined evaluation strategy integrates scalable quantitative analysis with expert clinical judgment, providing a comprehensive assessment of report-generation performance.
Radiologists were also invited to provide optional case-specific comments, which were subsequently analyzed to better understand the limitations of the current \ourAG system (\textbf{Supplementary Fig.~\ref{fig:supp_manual_feedback}}).
Reviewers were blinded to the model identity and study design during evaluation to minimize bias. Report quality was primarily assessed using RADPEER scores, with analysis focused on in-scope cases for which sufficient labeled supervision was available.

\subsection*{Details of competing methods}

On condition prediction tasks, key-slice selection tasks and segmentation tasks, we compare {\ourFM} to the following baseline models:

\begin{itemize}
    \item DINOv3: DINOv3 is a scalable vision foundation model trained with self-supervised learning, introducing Gram anchoring to preserve dense features, and delivering versatile, high-quality representations that outperform specialized models across diverse vision tasks without fine-tuning~\cite{simeoni2025dinov3}. DINOv3 extends DINOv2's~\cite{oquab2024dinov2} strengths in self-supervised ViT pretraining with better scalability, stability, dense feature quality, and adaptability, making it a more general and powerful vision foundation model.
    \item BrainMVP: BrainMVP is a multi-view vision pre-training framework for brain MRI that leverages $2.4$ million images across eight views (e.g., T1, T2, FLAIR) from diverse centers. Unlike uni-view models, it addresses missing views and fuses complementary information through cross-view reconstruction, view-level distillation, and view-aware contrastive learning. This enables the model to capture robust cross-view embeddings and generalize across clinical tasks~\cite{rui2024brainmvp}.
    \item BME-X: BME-X is a brain MRI enhancement foundation model designed for motion correction, super-resolution, denoising, and harmonization, improving both image quality and downstream analyses. It combines a tissue-classification network with a tissue-aware enhancement network to generate high-quality MR images.~\cite{sun2025foundation}.
    \item BiomedGPT: BiomedGPT is an open-source, lightweight vision-language foundation model designed as a biomedical generalist, capable of handling diverse tasks such as visual question answering, report generation, and summarization. Unlike traditional task-specific or closed-source systems, BiomedGPT offers flexibility and accessibility while maintaining a computing-friendly scale~\cite{zhang2024generalist,peng2025scaling}.
    \item ResNet50: ResNet is a residual learning framework that enables training of very deep neural networks by reformulating layers to learn residual functions relative to their inputs. Designed for image recognition, ResNets are easier to optimize. Among its variants, ResNet50 has become one of the most widely adopted architectures, achieving a balance between depth, efficiency, and performance. Despite the rise of larger vision transformers and foundation models, ResNet50 remains a state-of-the-art baseline due to its robustness, computational efficiency, and consistent performance across diverse domains and transfer learning tasks~\cite{he2016deep}.
\end{itemize}

On multimodal retrieval tasks, in addition to BiomedGPT, which is also used as a baseline in other evaluation settings, we compare \ourFM against a set of representative vision–language foundation models that are widely used for image–text alignment and cross-modal retrieval:
\begin{itemize}
    \item BiomedCLIP: BiomedCLIP is a domain-adapted variant of CLIP trained on large-scale biomedical image–text pairs, spanning radiology, pathology, and clinical literature. By aligning visual and textual representations within the biomedical domain, BiomedCLIP provides strong zero-shot performance on medical image–text retrieval tasks and serves as a widely adopted baseline for biomedical multimodal learning~\cite{zhang2023biomedclip}.
    \item MedSigLIP: MedSigLIP is a medical vision–language model derived from the SigLIP framework and pretrained on curated medical image–text data. It emphasizes improved scaling behavior and stable contrastive alignment, enabling robust cross-modal retrieval performance across diverse clinical imaging modalities~\cite{sellergren2025medgemma}.
    \item CLIP: CLIP is a general-purpose vision–language foundation model trained on large-scale natural image–text pairs using contrastive learning. Although not specifically optimized for medical imaging, CLIP remains a strong and widely used baseline for evaluating multimodal representation learning and zero-shot retrieval performance~\cite{radford2021learning}.
\end{itemize}

On the medical report generation task, we compare {\ourAG} against a representative vision–language baseline that directly maps medical images to free-text reports without explicit intermediate reasoning:

\begin{itemize}
    \item LLaVA: LLaVA is a general-purpose vision–language model that combines a pretrained visual encoder with a large language model to generate textual outputs conditioned on images. In our setting, we adapt LLaVA to accept spine MRI inputs and radiology-specific prompts, enabling end-to-end image-to-report generation. This baseline represents a black-box multimodal generation paradigm and serves as a strong reference for assessing the benefits of {\ourAG}'s structured, multi-agent report drafting framework~\cite{liu2023visual}.
\end{itemize}

\section*{Code and data availability}

The code has been provided as supplementary material and is available in a public repository at \url{https://github.com/qingfengtommy/SpineAgent}. The underlying imaging data are not publicly available owing to patient privacy considerations and institutional regulations.


\bibliographystyle{naturemag}
\bibliography{references}

@article{vos2020global,
  title={Global burden of 369 diseases and injuries in 204 countries and territories, 1990--2019: a systematic analysis for the Global Burden of Disease Study 2019},
  author={Vos, Theo and Lim, Stephen S and Abbafati, Cristiana and Abbas, Kaja M and Abbasi, Mohammad and Abbasifard, Mitra and Abbasi-Kangevari, Mohsen and Abbastabar, Hedayat and Abd-Allah, Foad and Abdelalim, Ahmed and others},
  journal={The lancet},
  volume={396},
  number={10258},
  pages={1204--1222},
  year={2020},
  publisher={Elsevier}
}

@article{lu2025global,
  title={Global, regional, and national burden of spinal cord injury from 1990 to 2021 and projections for 2050: A systematic analysis for the Global Burden of Disease 2021 study},
  author={Lu, Yubao and Shang, Zhizhong and Zhang, Wei and Hu, Xuchang and Shen, Ruoqi and Zhang, Keni and Zhang, Yuxin and Zhang, Liangming and Liu, Bin and Pang, Mao and others},
  journal={Ageing Research Reviews},
  volume={103},
  pages={102598},
  year={2025},
  publisher={Elsevier}
}

@article{hartvigsen2018low,
  title={What low back pain is and why we need to pay attention},
  author={Hartvigsen, Jan and Hancock, Mark J and Kongsted, Alice and Louw, Quinette and Ferreira, Manuela L and Genevay, St{\'e}phane and Hoy, Damian and Karppinen, Jaro and Pransky, Glenn and Sieper, Joachim and others},
  journal={The Lancet},
  volume={391},
  number={10137},
  pages={2356--2367},
  year={2018},
  publisher={Elsevier}
}

@article{GlobalEpidemicLowBackPain2023,
  title        = {The global epidemic of low back pain},
  author       = {{The Lancet Rheumatology}},
  journal      = {The Lancet Rheumatology},
  volume       = {5},
  number       = {6},
  pages        = {e305},
  year         = {2023},
  doi          = {10.1016/S2665-9913(23)00133-9},
  url          = {https://www.thelancet.com/journals/lanrhe/article/PIIS2665-9913(23)00133-9/fulltext}
}

@article{afshari2025growing,
  title={The growing nationwide radiologist shortage: current opportunities and ongoing challenges for international medical graduate radiologists},
  author={Afshari Mirak, Sohrab and Tirumani, Sree Harsha and Ramaiya, Nikhil and Mohamed, Inas},
  journal={Radiology},
  volume={314},
  number={3},
  pages={e232625},
  year={2025},
  publisher={Radiological Society of North America}
}

@book{hashemi2010mri,
  title={MRI: the basics},
  author={Hashemi, Ray H and Bradley, William G and Lisanti, Christopher J},
  year={2010},
  publisher={Lippincott Williams \& Wilkins}
}

@article{wu2025vision,
  title={Vision-language foundation model for 3D medical imaging},
  author={Wu, Jing and Wang, Yuli and Zhong, Zhusi and Liao, Weihua and Trayanova, Natalia and Jiao, Zhicheng and Bai, Harrison X},
  journal={npj Artificial Intelligence},
  volume={1},
  number={1},
  pages={17},
  year={2025},
  publisher={Nature Publishing Group UK London}
}

@article{bluethgen2025vision,
  title={A vision--language foundation model for the generation of realistic chest x-ray images},
  author={Bluethgen, Christian and Chambon, Pierre and Delbrouck, Jean-Benoit and Van Der Sluijs, Rogier and Po{\l}acin, Ma{\l}gorzata and Zambrano Chaves, Juan Manuel and Abraham, Tanishq Mathew and Purohit, Shivanshu and Langlotz, Curtis P and Chaudhari, Akshay S},
  journal={Nature Biomedical Engineering},
  volume={9},
  number={4},
  pages={494--506},
  year={2025},
  publisher={Nature Publishing Group UK London}
}

@article{xu2024whole,
  title={A whole-slide foundation model for digital pathology from real-world data},
  author={Xu, Hanwen and Usuyama, Naoto and Bagga, Jaspreet and Zhang, Sheng and Rao, Rajesh and Naumann, Tristan and Wong, Cliff and Gero, Zelalem and Gonz{\'a}lez, Javier and Gu, Yu and others},
  journal={Nature},
  volume={630},
  number={8015},
  pages={181--188},
  year={2024},
  publisher={Nature Publishing Group UK London}
}

@article{forsth2016randomized,
  title={A randomized, controlled trial of fusion surgery for lumbar spinal stenosis},
  author={F{\"o}rsth, Peter and {\'O}lafsson, Gylfi and Carlsson, Thomas and Frost, Anders and Borgstr{\"o}m, Fredrik and Fritzell, Peter and {\"O}hagen, Patrik and Micha{\"e}lsson, Karl and Sand{\'e}n, Bengt},
  journal={New England Journal of Medicine},
  volume={374},
  number={15},
  pages={1413--1423},
  year={2016},
  publisher={Mass Medical Soc}
}

@article{milligan2019degenerative,
  title={Degenerative cervical myelopathy: diagnosis and management in primary care},
  author={Milligan, James and Ryan, Kayla and Fehlings, Michael and Bauman, Craig},
  journal={Canadian Family Physician},
  volume={65},
  number={9},
  pages={619--624},
  year={2019},
  publisher={The College of Family Physicians of Canada}
}

@article{katz2022diagnosis,
  title={Diagnosis and management of lumbar spinal stenosis: a review},
  author={Katz, Jeffrey N and Zimmerman, Zoe E and Mass, Hanna and Makhni, Melvin C},
  journal={Jama},
  volume={327},
  number={17},
  pages={1688--1699},
  year={2022},
  publisher={American Medical Association}
}

@article{simeoni2025dinov3,
  title={{DINOv3}},
  author={Sim{\'e}oni, Oriane and Vo, Huy V and Seitzer, Maximilian and Baldassarre, Federico and Oquab, Maxime and Jose, Cijo and Khalidov, Vasil and Szafraniec, Marc and Yi, Seungeun and Ramamonjisoa, Micha{\"e}l and others},
  journal={arXiv preprint arXiv:2508.10104},
  year={2025}
}

@inproceedings{darcet2023vision,
  title={Vision Transformers Need Registers},
  author={Darcet, Timoth{\'e}e and Oquab, Maxime and Mairal, Julien and Bojanowski, Piotr},
  booktitle={The Twelfth International Conference on Learning Representations},
  year={2023}
}

@inproceedings{dosovitskiyimage,
  title={{An Image is Worth 16x16 Words: Transformers for Image Recognition at Scale}},
  author={Dosovitskiy, Alexey and Beyer, Lucas and Kolesnikov, Alexander and Weissenborn, Dirk and Zhai, Xiaohua and Unterthiner, Thomas and Dehghani, Mostafa and Minderer, Matthias and Heigold, Georg and Gelly, Sylvain and others},
  booktitle={International Conference on Learning Representations},
  year={2021}
}

@inproceedings{radford2021learning,
  title={{Learning Transferable Visual Models From Natural Language Supervision}},
  author={Radford, Alec and Kim, Jong Wook and Hallacy, Chris and Ramesh, Aditya and Goh, Gabriel and Agarwal, Sandhini and Sastry, Girish and Askell, Amanda and Mishkin, Pamela and Clark, Jack and others},
  booktitle={International conference on machine learning},
  pages={8748--8763},
  year={2021},
  organization={PmLR}
}

@article{oquab2024dinov2,
  title={{DINOv2: Learning Robust Visual Features without Supervision}},
  author={Oquab, Maxime and Darcet, Timoth{\'e}e and Moutakanni, Th{\'e}o and Vo, Huy and Szafraniec, Marc and Khalidov, Vasil and Fernandez, Pierre and Haziza, Daniel and Massa, Francisco and El-Nouby, Alaaeldin and others},
  journal={Transactions on Machine Learning Research Journal},
  pages={1--31},
  year={2024}
}

@article{rui2024brainmvp,
  title={{BrainMVP: Multi-modal Vision Pre-training for Brain Image Analysis using Multi-parametric MRI}},
  author={Rui, Shaohao and Chen, Lingzhi and Tang, Zhenyu and Wang, Lilong and Liu, Mianxin and Zhang, Shaoting and Wang, Xiaosong},
  journal={arXiv e-prints},
  pages={arXiv--2410},
  year={2024}
}

@article{sun2025foundation,
  title={{A foundation model for enhancing magnetic resonance images and downstream segmentation, registration and diagnostic tasks}},
  author={Sun, Yue and Wang, Limei and Li, Gang and Lin, Weili and Wang, Li},
  journal={Nature Biomedical Engineering},
  volume={9},
  number={4},
  pages={521--538},
  year={2025},
  publisher={Nature Publishing Group UK London}
}

@article{zhang2024generalist,
  title={{A generalist vision--language foundation model for diverse biomedical tasks}},
  author={Zhang, Kai and Zhou, Rong and Adhikarla, Eashan and Yan, Zhiling and Liu, Yixin and Yu, Jun and Liu, Zhengliang and Chen, Xun and Davison, Brian D and Ren, Hui and others},
  journal={Nature Medicine},
  pages={1--13},
  year={2024},
  publisher={Nature Publishing Group US New York}
}

@article{peng2025scaling,
  title={{Scaling Up Biomedical Vision-Language Models: Fine-Tuning, Instruction Tuning, and Multi-Modal Learning}},
  author={Peng, Cheng and Zhang, Kai and Lyu, Mengxian and Liu, Hongfang and Sun, Lichao and Wu, Yonghui},
  journal={arXiv preprint arXiv:2505.17436},
  year={2025}
}

@inproceedings{he2016deep,
  title={Deep residual learning for image recognition},
  author={He, Kaiming and Zhang, Xiangyu and Ren, Shaoqing and Sun, Jian},
  booktitle={Proceedings of the IEEE conference on computer vision and pattern recognition},
  pages={770--778},
  year={2016}
}

@article{richards2025rsna,
  title={The RSNA Lumbar Degenerative Imaging Spine Classification (LumbarDISC) Dataset},
  author={Richards, Tyler J and Flanders, Adam E and Colak, Errol and Prevedello, Luciano M and Ball, Robyn L and Kitamura, Felipe and Mongan, John and Vazirabad, Maryam and Lin, Hui-Ming and Kendell, Anne and others},
  journal={arXiv preprint arXiv:2506.09162},
  year={2025}
}

@article{lee2024performance,
  title={Performance of the winning algorithms of the RSNA 2022 cervical spine fracture detection challenge},
  author={Lee, Ghee Rye and Flanders, Adam E and Richards, Tyler and Kitamura, Felipe and Colak, Errol and Lin, Hui Ming and Ball, Robyn L and Talbott, Jason and Prevedello, Luciano M},
  journal={Radiology: Artificial Intelligence},
  volume={6},
  number={1},
  pages={e230256},
  year={2024},
  publisher={Radiological Society of North America}
}

@article{stephens2024rsna,
  title={Rsna launches lumbar spine degenerative classification ai challenge},
  author={Stephens, Keri},
  journal={AXIS Imaging News},
  year={2024},
  publisher={Anthem Media Group}
}

@inproceedings{chakraborty2020biomedbert,
  title={BioMedBERT: A Pre-trained Biomedical Language Model for QA and IR},
  author={Chakraborty, Souradip and Bisong, Ekaba and Bhatt, Shweta and Wagner, Thomas and Elliott, Riley and Mosconi, Francesco},
  booktitle={Proceedings of the 28th international conference on computational linguistics},
  pages={669--679},
  year={2020}
}

@article{liu2023visual,
  title={{Visual Instruction Tuning}},
  author={Liu, Haotian and Li, Chunyuan and Wu, Qingyang and Lee, Yong Jae},
  journal={Advances in neural information processing systems},
  volume={36},
  pages={34892--34916},
  year={2023}
}

@article{zhang2023biomedclip,
  title={{BiomedCLIP: a multimodal biomedical foundation model pretrained from fifteen million scientific image-text pairs}},
  author={Zhang, Sheng and Xu, Yanbo and Usuyama, Naoto and Xu, Hanwen and Bagga, Jaspreet and Tinn, Robert and Preston, Sam and Rao, Rajesh and Wei, Mu and Valluri, Naveen and others},
  journal={arXiv preprint arXiv:2303.00915},
  year={2023}
}

@inproceedings{zhai2023sigmoid,
  title={Sigmoid loss for language image pre-training},
  author={Zhai, Xiaohua and Mustafa, Basil and Kolesnikov, Alexander and Beyer, Lucas},
  booktitle={Proceedings of the IEEE/CVF international conference on computer vision},
  pages={11975--11986},
  year={2023}
}

@article{sellergren2025medgemma,
  title={{MedGemma Technical Report}},
  author={Sellergren, Andrew and Kazemzadeh, Sahar and Jaroensri, Tiam and Kiraly, Atilla and Traverse, Madeleine and Kohlberger, Timo and Xu, Shawn and Jamil, Fayaz and Hughes, C{\'\i}an and Lau, Charles and others},
  journal={arXiv preprint arXiv:2507.05201},
  year={2025}
}

@article{chowa2026language,
  title={From language to action: a review of large language models as autonomous agents and tool users},
  author={Chowa, Sadia Sultana and Alvi, Riasad and Rahman, Subhey Sadi and Rahman, Md Abdur and Raiaan, Mohaimenul Azam Khan and Islam, Md Rafiqul and Hussain, Mukhtar and Azam, Sami},
  journal={Artificial Intelligence Review},
  year={2026},
  publisher={Springer}
}

@inproceedings{shangagentsquare,
  title={AgentSquare: Automatic LLM Agent Search in Modular Design Space},
  author={Shang, Yu and Li, Yu and Zhao, Keyu and Ma, Likai and Liu, Jiahe and Xu, Fengli and Li, Yong},
  booktitle={The Thirteenth International Conference on Learning Representations},
  year={2025}
}

@article{fedus2022switch,
  title={Switch transformers: Scaling to trillion parameter models with simple and efficient sparsity},
  author={Fedus, William and Zoph, Barret and Shazeer, Noam},
  journal={Journal of Machine Learning Research},
  volume={23},
  number={120},
  pages={1--39},
  year={2022}
}

@inproceedings{rosenbaum2018routing,
  title={Routing Networks: Adaptive Selection of Non-Linear Functions for Multi-Task Learning},
  author={Rosenbaum, Clemens and Klinger, Tim and Riemer, Matthew},
  booktitle={International Conference on Learning Representations},
  year={2018}
}

@inproceedings{chen2020generative,
  title={Generative pretraining from pixels},
  author={Chen, Mark and Radford, Alec and Child, Rewon and Wu, Jeffrey and Jun, Heewoo and Luan, David and Sutskever, Ilya},
  booktitle={International conference on machine learning},
  pages={1691--1703},
  year={2020},
  organization={PMLR}
}

@article{abdin2024phi,
  title={Phi-4 Technical Report},
  author={Abdin, Marah I and Aneja, Jyoti and Behl, Harkirat S and Bubeck, S{\'e}bastien and Eldan, Ronen and Gunasekar, Suriya and Harrison, Michael and Hewett, Russell J and Javaheripi, Mojan and Kauffmann, Piero and others},
  journal={CoRR},
  year={2024}
}

@article{grattafiori2024llama,
  title={The llama 3 herd of models},
  author={Grattafiori, Aaron and Dubey, Abhimanyu and Jauhri, Abhinav and Pandey, Abhinav and Kadian, Abhishek and Al-Dahle, Ahmad and Letman, Aiesha and Mathur, Akhil and Schelten, Alan and Vaughan, Alex and others},
  journal={arXiv preprint arXiv:2407.21783},
  year={2024}
}

@article{hu2022lora,
  title={Lora: Low-rank adaptation of large language models.},
  author={Hu, Edward J and Shen, Yelong and Wallis, Phillip and Allen-Zhu, Zeyuan and Li, Yuanzhi and Wang, Shean and Wang, Lu and Chen, Weizhu and others},
  journal={ICLR},
  volume={1},
  number={2},
  pages={3},
  year={2022}
}

@article{goldberg2017acr,
  title={ACR RADPEER committee white paper with 2016 updates: revised scoring system, new classifications, self-review, and subspecialized reports},
  author={Goldberg-Stein, Shlomit and Frigini, L Alexandre and Long, Scott and Metwalli, Zeyad and Nguyen, Xuan V and Parker, Mark and Abujudeh, Hani},
  journal={Journal of the American College of Radiology},
  volume={14},
  number={8},
  pages={1080--1086},
  year={2017},
  publisher={Elsevier}
}

@article{brandao2013comparing,
  title={Comparing T1-weighted and T2-weighted three-point Dixon technique with conventional T1-weighted fat-saturation and short-tau inversion recovery (STIR) techniques for the study of the lumbar spine in a short-bore MRI machine},
  author={Brand{\~a}o, S and Seixas, D and Ayres-Basto, M and Castro, S and Neto, J and Martins, C and Ferreira, JC and Parada, F},
  journal={Clinical radiology},
  volume={68},
  number={11},
  pages={e617--e623},
  year={2013},
  publisher={Elsevier}
}

@article{monshi2020deep,
  title={Deep learning in generating radiology reports: A survey},
  author={Monshi, Maram Mahmoud A and Poon, Josiah and Chung, Vera},
  journal={Artificial Intelligence in Medicine},
  volume={106},
  pages={101878},
  year={2020}
}

@article{mcdermott2024closer,
  title={A closer look at auroc and auprc under class imbalance},
  author={McDermott, Matthew and Zhang, Haoran and Hansen, Lasse and Angelotti, Giovanni and Gallifant, Jack},
  journal={Advances in Neural Information Processing Systems},
  volume={37},
  pages={44102--44163},
  year={2024}
}

@article{ziegler2020open,
  title={Open health imaging foundation viewer: an extensible open-source framework for building web-based imaging applications to support cancer research},
  author={Ziegler, Erik and Urban, Trinity and Brown, Danny and Petts, James and Pieper, Steve D and Lewis, Rob and Hafey, Chris and Harris, Gordon J},
  journal={JCO clinical cancer informatics},
  volume={4},
  pages={336--345},
  year={2020},
  publisher={American Society of Clinical Oncology}
}
\clearpage
\section*{Supplementary Information}

\subsection*{Supplementary Prompts}


\begin{promptbox}{System Prompt}
system_prompt = """(*\beginToken*)(*\headerToken{system}*).
You are Spine-Agent, a radiology assistant focused strictly on Spine MRI.

Scope and inputs:
- Only discuss Spine MRI and directly related medical context.
- You may be given T1-weighted, T2-weighted, or both.
- Use auxiliary data as supportive context; always validate against the images.

Primary tasks:

1) Report Generation
- Produce a structured radiology report with professional tone.
- Summarize key actionable findings in the Impression.

Style and safety:
- Keep responses clear, concise, and focused on the provided Spine MRI.
- Do not fabricate unseen sequences, planes, or prior comparisons.
- Add an appropriate medical caution when relevant."""
\end{promptbox}

\vspace{1em} 

\begin{promptbox}{Report Generation Prompt}
agent_report_generation_prompt = """(*\imageToken*)
If provided, you may also receive auxiliary data delimited by special tokens:
- Disease condition prediction results inside <diagnosis> and </diagnosis>. 0 represents absence of a disease type, 1 represents presence of a disease type.
- Top-1 similar case report inside <similar_case_report> and </similar_case_report>

Use auxiliary data as supportive context only. Always prioritize image evidence; if there is a conflict, state the discrepancy clearly. Do not copy text verbatim from the similar report

<diagnosis>{{}}</diagnosis> <similar_case_report>{{}}</similar_case_report> Please analyze these MRI scans and generate a comprehensive radiology report including all findings, measurements, and clinical observations.

The templates below are the standard templates for reporting MRI of the cervical, thoracic, and lumbar spine. If intravenous contrast was administered, enhancement is usually described in the subcomponents of the template.

[LUMBAR SPINE TEMPLATE]
FINDINGS:
ALIGNMENT: Normal - Description of the alignment of vertebra in the l-spine.
MARROW: Normal - Description of the bone marrow and any lesions in the vertebra.
DISCS: Discs are normal in height and signal intensity. - Description of the inter-vertebral discs, particularly their height loss and desiccation or abnormal signal.
CORD: Conus ends normally at L1-L2. Visualized cord and cauda equina are normal. - Description of the conus and cauda equina including intramedullary and intradural extramedullary lesions.
PARAVERTEBRAL SOFT TISSUES: Normal - Description of findings that are outside of the vertebra and spinal canal in the visualized organs and soft tissues.
AXIAL DISCS, DURAL COMPRESSION & FORAMINA: - The individual anatomic levels below are used to describe primarily the spinal canal, lateral recess, and neural foramen at each anatomic level and the degree of compromise as well as the etiology for that compression/narrowing.
L1-2: No central or foraminal stenosis. Facets are normal. 
L2-3: No central or foraminal stenosis. Facets are normal. 
L3-4: No central or foraminal stenosis. Facets are normal. 
L4-5: No central or foraminal stenosis. Facets are normal. 
L5-S1: No central or foraminal stenosis. Facets are normal. 
IMPRESSION: - The impression is a concise restating of the clinically important findings of the report as well as the interpretation of those findings which may include 
specific diagnoses. The impression is also used to describe important items that are not present but that need to be understood by the treating provider to determine the next course in clinical management. 
Provide these diagnoses or issues as an enumerated list.
1.
2. 

[THORACIC SPINE TEMPLATE]
FINDINGS:
ALIGNMENT: Normal - Description of the alignment of vertebra in the l-spine.
MARROW: Normal - Description of the bone marrow and any lesions in the vertebra.
DISCS: Discs are normal in height and signal intensity. - Description of the inter-vertebral discs, particularly their height loss and desiccation or abnormal signal.
CORD: Visualized spinal cord is normal in signal and size. - Description of the conus and cauda equina including intramedullary and intradural extramedullary lesions.
PARAVERTEBRAL SOFT TISSUES: Normal - Description of findings that are outside of the vertebra and spinal canal in the visualized organs and soft tissues.
SPECIFIC LEVELS: Abnormal levels are described separately below. - Since pathology in the thoracic spine is less common the individual anatomic vertebral levels are usually not 
enumerated. Use this section to describe the specific levels that are pathologic, particularly in terms of: spinal canal, lateral recess, and neural foramenal compromise.
ALL OTHER LEVELS: No central or foraminal stenosis. Facets are normal.
IMPRESSION: - The impression is a concise restating of the clinically important findings of the report as well as the interpretation of those findings which may include 
specific diagnoses. The impression is also used to describe important items that are not present but that need to be understood by the treating provider to determine the next course in clinical management. 
Provide these diagnoses or issues as an enumerated list.
1.
2. 

[CERVICAL SPINE TEMPLATE]
FINDINGS:
ALIGNMENT: Normal - Description of the alignment of vertebra in the l-spine.
MARROW: Normal - Description of the bone marrow and any lesions in the vertebra.
DISCS: Discs are normal in height and signal intensity. - Description of the inter-vertebral discs, particularly their height loss and desiccation or abnormal signal.
CORD: Visualized spinal cord is normal in signal and size. - Description of the conus and cauda equina including intramedullary and intradural extramedullary lesions.
PARAVERTEBRAL SOFT TISSUES: Normal - Description of findings that are outside of the vertebra and spinal canal in the visualized organs and soft tissues.
AXIAL DISCS, DURAL COMPRESSION & FORAMINA: - The individual anatomic levels below are used to describe primarily the spinal canal, lateral recess, and neural foramen at each anatomic level and the degree of compromise as well as the etiology for that compression/narrowing.
C2-3: No central or foraminal stenosis. Facets are normal. 
C3-4: No central or foraminal stenosis. Facets are normal. 
C4-5: No central or foraminal stenosis. Facets are normal. 
C5-6: No central or foraminal stenosis. Facets are normal. 
C6-7: No central or foraminal stenosis. Facets are normal. 
C7-T1: No central or foraminal stenosis. Facets are normal. 
IMPRESSION: - The impression is a concise restating of the clinically important findings of the report as well as the interpretation of those findings which may include specific diagnoses. The impression is also used to describe important items that are not present but that need to be understood by the treating provider to determine the next course in clinical management. 
Provide these diagnoses or issues as an enumerated list.
1. 
2.

Please first identify whether the MRI scans are of the cervical, thoracic, or lumbar spine, and then use the corresponding template to generate the report. If the classification result suggests a region, still verify on the images.
<report_generation>"""
\end{promptbox}

\begin{inputnote}
The \imageToken{} tag represents a multi-modal input payload consisting of:
\begin{itemize}
    \item \textbf{Primary MRI Series:} Full-stack T1/T2/others sequences.
    \item \textbf{Segmentation Result:} High-resolution crops of "key-regions" identified by the Segmentation Agent to ensure focus on pathologic levels.
\end{itemize}
\end{inputnote}

\subsection*{Supplementary Figures}

\begin{figure}[h]
    \centering
    \includegraphics[width=1\linewidth]{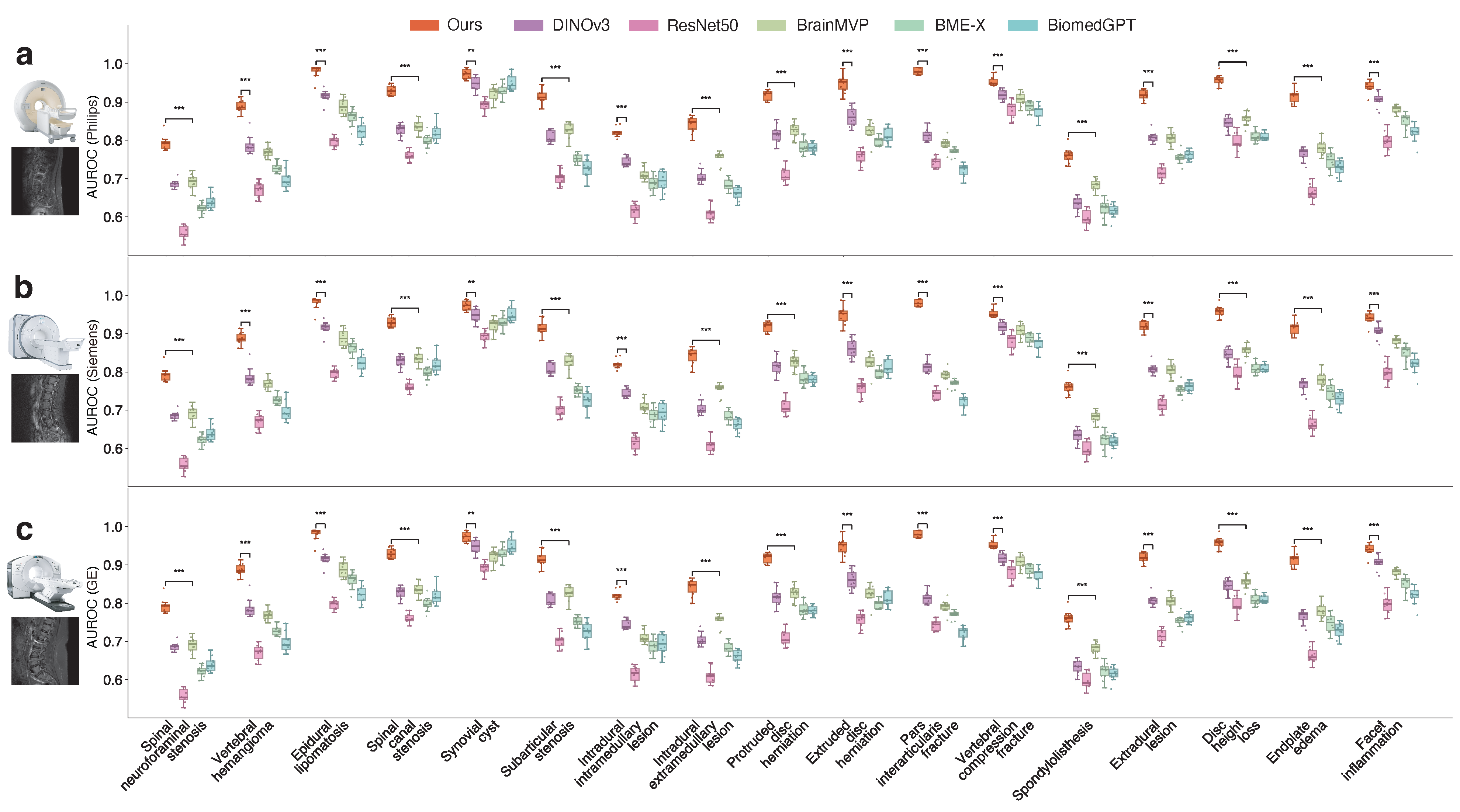}
    \suppfigurecaption{\textbf{Cross-manufacturer performance} of \ourAG and baseline methods measured by their AUROC scores at the individual condition level. \ourAG consistently outperforms competing methods, demonstrating the robustness and generalization benefits of the pretrained foundation model \ourFM across heterogeneous scanner manufacturers.}
    \label{fig:supp_crossmanufacturer}
\end{figure}

\end{document}